\documentclass{article} 
\usepackage[preprint]{colm2026_conference}

\usepackage{microtype}
\usepackage[T1]{fontenc}
\usepackage{hyperref}
\usepackage{url}
\usepackage{booktabs}

\usepackage{lineno}

\definecolor{darkblue}{rgb}{0, 0, 0.5}
\hypersetup{colorlinks=true, citecolor=darkblue, linkcolor=darkblue, urlcolor=darkblue}

\usepackage{graphicx}
\usepackage{booktabs}
\usepackage{wrapfig}
\usepackage{multirow}
\usepackage{amsfonts}
\usepackage{amsmath}
\usepackage{nicefrac}
\usepackage{algorithm}
\usepackage{algpseudocode}

\usepackage{adjustbox}

\usepackage{listings}
\usepackage[most]{tcolorbox}

\lstdefinelanguage{none}{}
\DeclareTextCommandDefault{\textquotedbl}{\char34}

\newtcblisting{promptbox}[1][]{
  enhanced,
  colback=white,
  colframe=black,
  boxrule=0.5pt,
  arc=2pt,
  left=8pt,
  right=8pt,
  top=6pt,
  bottom=6pt,
  colbacktitle=gray!10,
  coltitle=black,
  fonttitle=\bfseries\small,
  title={#1},
  listing only,
	  listing options={
	    language=none,
	    basicstyle=\ttfamily\small,
	    columns=fullflexible,
	    breaklines=true,
	    breakatwhitespace=false,
	    breakautoindent=false,
	    breakindent=0pt,
	    showstringspaces=false,
	    escapeinside={(*@}{@*)},
	    literate=
      {"}{{\char34}}1
      {’}{{'}}1
      {“}{{\char34}}1
      {”}{{\char34}}1
      {—}{{\textemdash}}1
      {°}{{$^\circ$}}1
      {Â£}{{GBP}}1
      {ä}{{\"a}}1
      {ö}{{\"o}}1
      {ü}{{\"u}}1
      {Ä}{{\"A}}1
      {Ö}{{\"O}}1
      {Ü}{{\"U}}1
      {ß}{{\ss}}1
  }
}

\newtcblisting{artifactbox}[1][]{
  enhanced,
  breakable,
  width=\linewidth,
  colback=white,
  colframe=black,
  boxrule=0.5pt,
  arc=2pt,
  left=6pt,
  right=6pt,
  top=5pt,
  bottom=5pt,
  colbacktitle=gray!10,
  coltitle=black,
  fonttitle=\bfseries\small,
  title={#1},
  listing only,
  listing options={
    language=none,
    basicstyle=\ttfamily\footnotesize,
    columns=fullflexible,
    breaklines=true,
    breakatwhitespace=false,
    breakautoindent=false,
    breakindent=0pt,
    showstringspaces=false,
    literate=
      {"}{{\char34}}1
      {’}{{'}}1
      {“}{{\char34}}1
      {”}{{\char34}}1
      {—}{{\textemdash}}1
  }
}

\newcommand{\subtle}[1]{{\tiny\color[gray]{0.4}#1}}

\title{Learning User Simulators with Turing Rewards}

\author{Yingshan Susan Wang$^{1*}$, Cedegao E. Zhang$^{1*}$, Linlu Qiu$^{1*}$, Zexue He$^{2}\thanks{Core contributors.}$\:\:, \\
\vspace{2mm}
\textbf{Pengyuan Li$^3$, Alex Pentland$^{1,2}$, Roger P. Levy$^1$, Yoon Kim$^1$} \\
  $^1$Massachusetts Institute of Technology, $^2$Stanford University, $^3$MIT-IBM Watson AI Lab    \vspace{1mm} \\
  \texttt{\{susanw26, cedzhang, linluqiu\}@mit.edu}, \texttt{zexueh@stanford.edu} 
}

\begin{document}

\ifcolmsubmission
\linenumbers
\fi


\maketitle
\thispagestyle{plain}

\begin{abstract}
Learning to simulate human users in interactive settings could advance the training of agent assistants, evaluation of personalization systems, research in the social sciences, and more. Existing approaches generally do so by training a large language model (LLM) to match a single ground truth response, either by maximizing the log probability or by using a similarity reward. We instead propose Turing-RL: a Turing-Test-based reinforcement learning approach for training user simulator models. Turing-RL uses a discriminative Turing reward with an LLM judge to score how indistinguishable a generated response is from the real user's given the user's history, and the user simulator LLM learns to produce responses indistinguishable from what the user could have said with such rewards. Across two different domains---conversational chat and Reddit forum discussion---we find that Turing-RL consistently outperforms baseline methods on both LLM and human evaluation metrics. Our study suggests that optimizing for indistinguishability, rather than response matching, is effective for learning user simulators.\footnote{Code is available at: \url{https://github.com/SusanWYS/turing-rl}}
\end{abstract}


\section{Introduction}
The currently dominant use cases of large language models (LLMs) involves having them act as helpful assistants to human users. However, a growing range of applications would benefit  from their playing  the opposite role---i.e.,  not the assistant, but the user.
Such user simulators could serve as foundational building blocks for social world models in AI agents \citep{rabinowitz2018machine, collins2024building}, training environments and testbeds for interactive systems~\citep{abdulhai2025consistently, mehri2025goal}, and proxies for studying human behaviors at scale~\citep{aher2023using, park2023generative, lu2025multiturn}.

Simulating an individual is however a fundamentally difficult task. What distinguishes one person from another resists easy categorization: two people with identical demographics can hold sharply different opinions~\citep{hwang2023aligning, santurkar2023opinions}, and individual preferences cannot be recovered from group-level labels alone~\citep{ kirk2024prism, jiang2024indievalue}. Recent work has begun addressing this challenge through purpose-built user language models~\citep{naous2025flipping}, latent user-state alignment~\citep{wu2026humanlm}, and log-probability maximization with chain of thought~\citep{gandhi2026dialogue}.
These approaches share a common assumption: the training signal is derived from matching a specific ground truth response, whether by scoring similarity against it with an LLM judge or by  maximizing its log-probability. However, the set of plausible responses to a given context is enormous, i.e., the same person in the same context could say many different things. An ideal user simulator model should therefore go beyond replicating the ground truth response and instead produce responses indistinguishable from what the user {could have said}.




\begin{figure*}[t]
    \centering
    \includegraphics[width=0.9\textwidth]{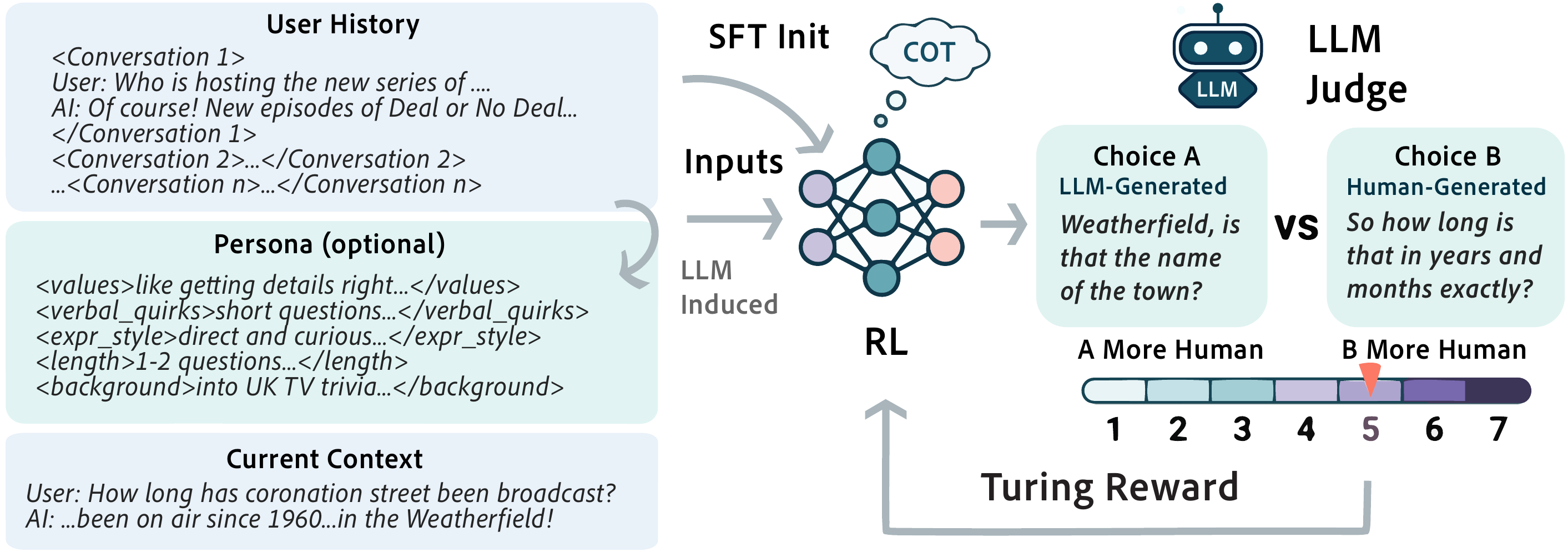}
    \caption{Overview of {Turing-RL}. Given a user's history, induced persona, and current conversation context, an SFT-initialized policy generates multiple candidate responses via chain-of-thought (CoT) reasoning. A Turing judge (LLM) compares each candidate against the ground truth human response on a 1--7 scale, scoring which is more likely written by the real user. This discriminative Turing reward is used to train the policy with GRPO.}
    \label{fig:main}
\end{figure*}

The classic Turing Test \citep{turing1950computing} operationalizes human-likeness through indistinguishability: a machine passes if an evaluator cannot tell its responses from a human's. This is precisely the criterion we want from a user simulator, and it translates directly into a training signal. In this paper, we therefore propose to train LLM user simulators via reinforcement learning (RL) on a \emph{discriminative Turing reward}, wherein an LLM judge assigns a high score when a generated response is deemed human-like, conditioned on the user's history, which includes the current session history and the user's behaviors in previous sessions. Figure~\ref{fig:main} shows an overview of our approach.

We evaluate our Turing-RL-trained LLMs across two settings that differ substantially in structure, in particular multi-turn dialogue~\citep{kirk2024prism} and Reddit forum discussions~\citep{chang2020convokit}. We compare against baseline training signals that have been previously explored in the literature: a response-similarity reward adapted from HumanLM~\citep{wu2026humanlm}, and log-probability maximization with chain-of-thought adapted from~\cite{gandhi2026dialogue}. We find that Turing-RL consistently outperforms both alternatives across both LLM and human evaluation metrics, suggesting that optimizing for  indistinguishability is an effective path towards learning user simulators.



\section{Learning User Simulators}

\subsection{Problem Formulation}\label{sec:problem}


We study user simulation in an interactive setting, where we are given the current session context $x$ (i.e., prior interactions in the current conversation session or thread) and a user representation $u$, and need to generate a response $y$ that could have been plausibly produced by the user. 

The most straightforward way to represent a user is through their behavior history $h$ (e.g., prior utterances or posts that are not part of the current session). For each user, we reserve a fixed block of $k$ prior interactions at dataset construction time and reuse it as $h$ across all prediction targets for that user. We require $h$ to be disjoint from the current target context $x$ and ground truth response $y_\star$, ensuring that the user representation is never contaminated by the target example. The size $k$ is sampled per user from a small range.

Additionally, we can also pair the raw behavior history with an induced persona $\rho$. We induce $\rho$ by prompting an auxiliary language model to summarize stable traits from the same user's history block $h$. See  Figure~\ref{fig:persona_induction} of the appendix for the prompt used for persona induction. We use $u = (h, \rho)$ as the default in most of our experiments, and ablate on the choice of $u$ (i.e., without the persona or without the history) in the ablation study.

\subsection{Turing-RL}\label{sec:training}
We use a discriminative Turing reward as the training signal. An LLM judge is shown the user's  history $(x, h)$ alongside two responses to the same context: one written by the real user from the ground truth distribution $y_{\star} \sim p_\star(\cdot | x,h)$ and one sampled from the model $y \sim p_\theta( \cdot | x, u )$. The judge rates on a $1$--$7$ Likert scale which response was written by the human, considering the local context, the user's motive, and stylistic fits. A score of $1$ means that the judge deems the human-generated response $y_{\star}$ to be more likely to be written by the human, while a score of $7$ means the judge deems the model-generated response $y$ is more likely to be written by the human.\footnote{In practice, we randomize the ordering of the responses given to the judge to mitigate against ordering bias. We then convert the randomized ordering to the standardized ones as described above.} Rather than optimizing for content overlap with a specific ground truth response, this signal trains the model toward the broader quality of being indistinguishable from the target user. The full judge prompt is given in Appendix~\ref{app:judge-prompts}.

Letting $s({y, y_\star}) \in \{1, \dots, 7\}$ be the score assigned by the LLM judge, we cap and  normalize the score to $[0,1]$ to obtain the Turing reward 
\begin{align*}
r_{\mathrm{turing}} (y, y_\star) =  \frac{\min\{s(y, y_{\star}), 5\} -1}{6}.
\end{align*}
The reward cap at $5$ mitigates against reward hacking: since we would like to produce responses that are human-like, a model that produces responses that are substantially ``more human'' (as deemed by the judge) than actual human responses would be undesirable and  be potentially indicative of reward hacking.\footnote{We indeed found this type of reward hacking to occur in practice in preliminary experiments.}
The final RL objective is then given by
\begin{align*}
    \max_\theta \,\, \mathbb{E}_{y \sim p_\theta(\cdot | x, u), y_\star \sim p_\star(\cdot | x, h)}\left[ r_{\text{turing}} (y, y_\star) \right],
\end{align*}
and we train with Group Relative Policy Optimization~\citep[GRPO;][]{shao2024deepseekmath}, after an initial supervised finetuning (SFT) phase on a disjoint subset of the training data.


\section{Experimental Setup}

\subsection{Datasets}
We evaluate our approach on two domains: multi-turn chat and Reddit forum discussion. Both domains contain interaction data annotated with user information. We reserve a fixed block of prior interactions as behavior history $h$, and use the remaining interactions as prediction targets. We split users into training and evaluation sets. Full preprocessing details and statistics are given in Appendix~\ref{app:dataset-stats}.


\paragraph{(Chat) Multi-turn chat: PRISM.}
The PRISM Alignment Dataset~\citep{kirk2024prism} contains multi-turn conversations between humans and LLM assistants, spanning 1,500 participants from 75 countries. We select users with at least 6 conversations (1,288 users) and hold out 128 users (880 target user-response turns) for evaluation. 

\paragraph{(Reddit) Online forum discussion: ConvoKit.}
The ConvoKit subreddit corpus~\citep{chang2020convokit} contains discussions on Reddit. We select 14 subreddits, selecting users with at least 8 threads (1,282 users), and hold out \texttt{r/tifu} and \texttt{r/worldnews} for evaluation (102 users, 267 examples), with no user overlap with training. Each target thread yields one example: the user's last comment, and the ancestor comment chain starting from the original post to the target comment.

\subsection{Training}
\paragraph{SFT warm start.}
Before RL, we warm-start the policy with SFT on ground truth user responses augmented with chain-of-thought reasoning traces, a common practice in LLM RL \cite[e.g.,][]{guo2025deepseek}. For each training example, we use the Qwen3-8B instruct model to generate a reasoning trace given the context and ground truth response, explaining what could have led the user to write that response. The SFT target pairs the trace with the ground truth: \texttt{<reasoning>}$t$\texttt{</reasoning> [HUMAN]:}~$y$. The model is trained with LoRA using completion-only loss. At inference time, the model generates both the trace and the response autoregressively. See Appendix~\ref{app:thinking-trace-prompt} for the trace elicitation prompt.

\paragraph{RL training.}
We optimize the Turing discriminative judge signal with GRPO~\citep{shao2024deepseekmath}. We sample 4 candidate responses per training example and compute advantages via within-group normalization. All models are initialized from the SFT checkpoint discussed above and trained with LoRA ($\text{rank }{r=}64$, $\text{scaling }\alpha{=}32$) for 3 epochs \citep{lora, qlora}. GRPO training uses a dataset disjoint from the SFT training set.
We  additionally apply \emph{length penalty term} that penalize responses falling outside a tolerance band around the ground truth length (see Appendix~\ref{app:length_penalty}). We use Qwen3.5-397B-A17B~\citep{qwenteam2026qwen35omni} as the judge. Training details are given in Appendix~\ref{app:training_details}. 

\subsection{Baselines}

We use Qwen3-8B with thinking mode disabled as the base model, from which we learn a user simulator model through our pipeline. We compare our method with two RL-based methods using alternative reward signals described below. To ensure fair comparison, all RL-based methods are initialized from the same SFT checkpoint, make use of the same history/personas, and are trained with GRPO. We also report the performance of the SFT-init checkpoint, base Qwen3-8B model, Qwen3.5-397B-A17B, and OpenAI GPT-5 as references.

\paragraph{Sim-RL: response similarity reward.}
Following \citet{wu2026humanlm}, this baseline trains with a reward that measures how well the generated response $y$ captures the content of the ground truth response $y_\star$. An LLM judge is given the user's history, current context, and both responses, and produces an overall alignment score $r_{\mathrm{sim}}(y, y_\star) \in [0,1]$ that reflects semantic similarity while penalizing unsupported content (see Appendix Figure~\ref{fig:sim_prompt} for prompt). This reward encourages the model to say roughly \emph{what} the user said by matching the key points in ground truths.

\paragraph{Logprob-RL: log-probability reward.}
\citet{gandhi2026dialogue} propose training dialogue simulators by using the log-probability of the ground truth response under the model as the reward, alongside chain-of-thought reasoning. The model generates a reasoning trace $z$ and a candidate response $y$, and the reward is given by
\begin{align*}
r_{\mathrm{logprob}}(z) = \frac{1}{|y_\star|}\log p_\theta(y_\star \mid x, u, z),
\end{align*}
where $|y_\star|$ is the number of tokens in the ground truth response and $z$ is the reasoning trace generated by the model. This reward roughly maximizes a lower bound on the log marginal likelihood of the ground truth response.\footnote{Without length normalization, Jensen's inequality gives $\mathbb{E}_z[\log p_\theta(y_\star \mid x,u,z)] \leq \log p_\theta(y_\star \mid x,u)$.}
In their full objective, this reward is combined with an auxiliary SFT term; in our implementation, we drop the auxiliary SFT loss term.\footnote{We found that dropping the SFT loss term was more stable for Qwen3-8B GRPO training.}

\subsection{Evaluation}
\subsubsection{LLM-as-a-Judge Evaluation}
We use LLM-as-a-judge to evaluate generated responses along three dimensions described below. Claude Sonnet 4.6~\citep{anthropic2026claudesonnet46} is used for all evaluations. See Appendix~\ref{app:judge-prompts} for the full judge prompts and Appendix~\ref{sec:sampling_params} for response generation configuration.

\paragraph{Turing distinguishability.}
 A judge is presented with the user's behavioral history, the interaction context, and two responses: the ground truth human response and a model-generated candidate, with position randomized. The judge rates which response was written by the real user on a 1--7 scale across three criteria (immediate target, human goal, and communication style). To control for position bias, we evaluate each pair in a randomized ordering and convert the ordering to a canonical one for scoring, fixing choice A to be the ground truth and choice B to be model-generated. Higher scores indicate the generated response is more human-like. This metric mirrors our training signal but makes use of a different and potentially more powerful\footnote{According to some benchmarks, e.g., \url{https://artificialanalysis.ai/leaderboards/models}.}   Sonnet 4.6 judge than the Qwen3.5-397B-A17B judge  used for training.

\paragraph{Response similarity to ground truth.} We also measure similarity between the generated and ground truth responses, following \citet{wu2026humanlm}.  An LLM judge scores how well a generated response captures the overall content of the ground truth human response. The judge extracts key points from the ground truth, then scores semantic similarity, penalizing extraneous content, wrong perspective, and source copying. This yields a single score in $[0, 1]$ measuring whether the model said roughly what the real user said; we report this score as a percentage in results.

\paragraph{Context and user specificity.}
We finally evaluate whether a generated response is specifically grounded in the current interaction and compatible with the target user, rather than being a generic but plausible reply, which is a common failure of conversational models \citep{li2016diversity, user_sim_survey}. The judge scores two dimensions: \emph{context specificity} (how tightly the response engages with the exact local interaction) and \emph{user evidence compatibility} (how compatible the response is with the target user's observed behavior without unnaturally exposing or inventing evidence), each weighted $0.5$. This yields a single score in $[0, 1]$. For better calibration, we batch judge the scores of multiple responses in one judge call.

\subsubsection{Human Evaluation}

In addition to LLM-as-a-judge evaluation, we also recruit 360 human participants from Prolific to perform a binary-choice Turing test: given a target user's history and two candidate responses (one real, one model-generated) in randomized order, annotators select which was written by the real user. We evaluate three models: LLM trained with just supervised finetuning (SFT-Init), RL-training with the similarity reward (Sim-RL), and RL-training with our Turing reward (Turing-RL). We do not test Logprob-RL for human evaluation as LLM-as-a-judge results indicate that this variant underperforms Sim-RL and Turing-RL. We target 100 heldout samples per dataset, collecting 600 binary judgments per condition (6 judgments for each target sample) after filtering for comprehension. We report the model win rate (WR), where WR ${>} 0.5$ means that the model response is picked by human annotators above chance. Full human evaluation details are given in Appendix~\ref{app:human-eval}.





\section{Results}

\subsection{LLM-as-a-Judge Evaluation}
\paragraph{Turing-RL trains  human-like simulators.}
\begin{figure}[t]
\centering
\includegraphics[width=\columnwidth]{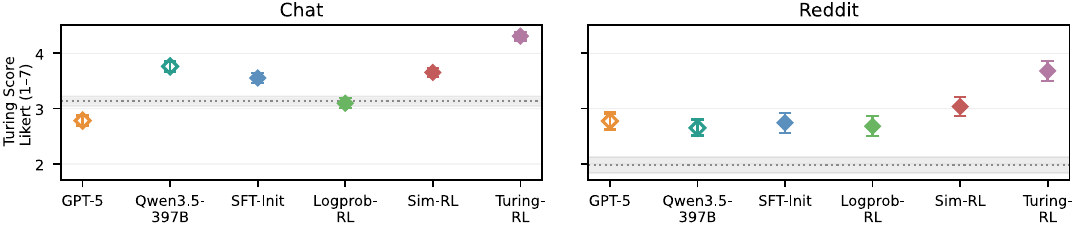}
\vspace{-2mm}
\caption{\textbf{LLMs trained with Turing rewards (Turing-RL) outperforms other training signals on human-likeness in both domains.} Turing judge scores (1--7 Likert; higher = more human-like) from Sonnet~4.6 on Chat and Reddit, with 95\% CIs. Hollow markers indicate untrained baselines, while solid markers denote trained Qwen3-8B variants. The dashed line with shaded band marks the Qwen3-8B base model performance with 95\% CI.}
\label{fig:turing_dot}
\end{figure}

\begin{figure}[t]
\vspace{2mm}
\centering
\includegraphics[width=\columnwidth]{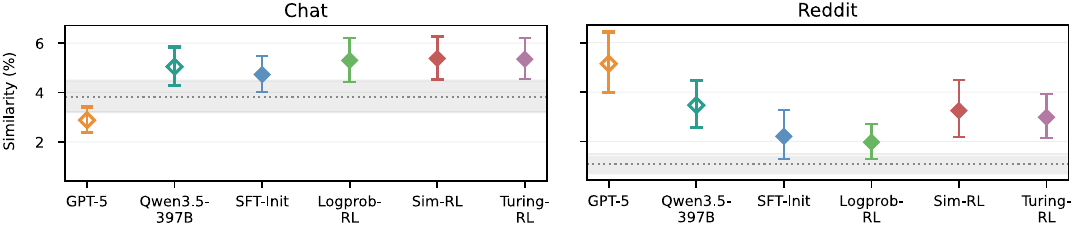}
\vspace{-2mm}
\caption{\textbf{Turing-RL matches Sim-RL even though Sim-RL is   explicitly trained to maximize similarity}, showing that optimizing for indistinguishability does not sacrifice content alignment. Response similarity to ground truth (Sim, \%; higher = more similar to what the user actually said) from Sonnet~4.6 on Chat and Reddit, with 95\% CIs. Hollow markers indicate untrained baselines, while solid markers denote trained Qwen3-8B variants. The dashed line with shaded band marks the Qwen3-8B base model performance with 95\% CI.}
\label{fig:sim_dot}
\end{figure}

Figure~\ref{fig:turing_dot} shows Turing judge scores across both domains. Turing-RL outperforms all other models, including Sim-RL and SFT-Init, on both Reddit and Chat. The gap is large on Chat, where Turing-RL exceeds the next-best model by a wide margin.

Notably, user simulation remains a difficult task overall. Even GPT-5 and Qwen3.5-397B---much more capable models than our user simulator---do not improve much compared to Qwen3-8B base. Qualitative inspection reveals that GPT-5 and Qwen3.5-397B tend to produce verbose, overly hedged responses that read as assistant-like rather than human-like (see Figure~\ref{fig:qualitative_examples_2} and Figure~\ref{fig:qualitative_examples}). Training with the similarity reward does not significantly improve the Turing score over the SFT-Init checkpoint, suggesting that matching the content of the ground truth response does not translate to human-like indistinguishability.

\paragraph{Turing-RL does not sacrifice similarity to ground truth.}
Figure~\ref{fig:sim_dot} reports similarity to the ground truth response. Among the trained models, Sim-RL and Turing-RL perform comparably, both improving over the SFT model. This confirms that training with the Turing reward does not sacrifice content alignment: Turing-RL produces responses that are both hard to distinguish from the real user and similar in content to what the user actually said. GPT-5 achieves notably high similarity on Reddit, likely because its verbose outputs happen to cover more of the ground truth key points, even though this verbosity hurts its Turing score. 

\paragraph{Turing-RL user simulators  are more grounded in context.}
\begin{figure}[t]
\centering
\includegraphics[width=\columnwidth]{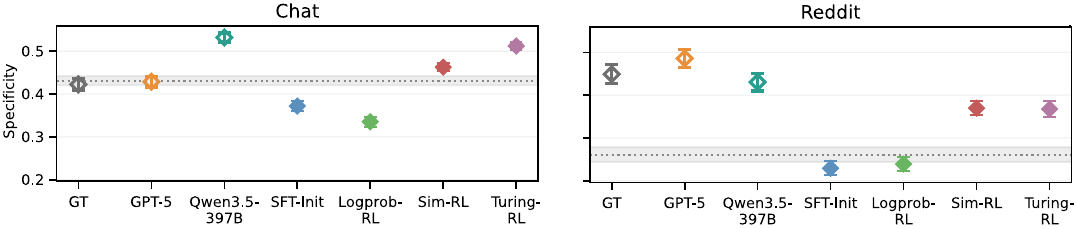}
\caption{\textbf{Turing-RL is among the strongest on Chat, while Turing-RL and Sim-RL are strongest among the trained models on Reddit},  outperforming SFT-Init and Logprob-RL. Response specificity ($[0,1]$; higher = more grounded in the interaction context and compatible with the target user) from Sonnet~4.6, with 95\% CIs. Hollow markers indicate untrained baselines, while solid markers denote trained Qwen3-8B variants. GT is the ground truth human response, and the dashed line marks the Qwen3-8B base model with 95\% CI.}
\label{fig:specificity}
\end{figure}

Beyond matching content, we measure whether generated responses are specifically grounded in the interaction context and compatible with the target user, rather than being generic but plausible replies. Figure~\ref{fig:specificity} shows that Turing-RL and Sim-RL outperform SFT-Init and Logprob-RL in both domains.
On Chat, Qwen3.5-397B scores highest overall, followed by Turing-RL, which exceeds both GT and GPT-5. Sim-RL also improves over SFT-Init. On Reddit, GPT-5, Qwen3.5-397B, and the ground truth (GT) score highest, while Turing-RL and Sim-RL are close among trained Qwen3-8B models. 

\subsection{Human Evaluation}

Table~\ref{tab:human_eval_winrate} reports mean win rates over ground truth responses. These estimates are computed from 100 heldout users per dataset (Reddit/Chat) with 6 selected annotators per heldout user, yielding 600 binary judgments per dataset--model pair.  On Chat, Turing-RL has the highest win rate (WR $= .57$), while SFT-Init and Sim-RL remain close to chance. On Reddit, both Sim-RL and Turing-RL are near chance and much higher than SFT-Init. 

\begin{wraptable}[9]{r}{0.4\textwidth}
\vspace{-3mm}
\centering
\small
\setlength{\tabcolsep}{0pt}
\begin{tabular*}{0.75\linewidth}{@{\extracolsep{\fill}}lcc@{}}
\toprule
 & \textbf{Chat} & \textbf{Reddit} \\
\midrule
SFT-Init    & .49\subtle{$\pm.061$} & .41\subtle{$\pm.045$} \\
Sim-RL    & .50\subtle{$\pm.055$} & .52\subtle{$\pm.049$} \\
Turing-RL & .57\subtle{$\pm.050$} & {.50\subtle{$\pm.051$}} \\
\bottomrule
\end{tabular*}
\vspace{-2mm}
\caption{Human Turing Test, model win rates against ground truth ($\pm$ 95\% CI).}
\vspace{4mm}
\label{tab:human_eval_winrate}
\end{wraptable}

We perform target-level paired permutation tests between Turing-RL and the other two models. On Chat, Turing-RL significantly outperforms both SFT-Init ($p=0.044$) and Sim-RL ($p=0.022$). On Reddit, Turing-RL significantly improves on human-likeness compared to SFT-Init ($p=0.0095$), and no statistically significant difference is detected between Turing-RL and Sim-RL. Further statistical test details are given in Appendix~\ref{app:human-eval-significance}. 

We note that Reddit is significantly harder for humans to judge than Chat. As shown in Tables~\ref{tab:human_eval_rt} and~\ref{tab:human_eval_rt_per_word}, the Chat/Reddit mean reaction time ratio per target question is 1.43, and the Reddit/Chat mean reaction time per word ratio is 1.48. Therefore, we should treat the human study as a validation of broad trends rather than conclusive evidence, especially for Reddit. Overall, humans agree with LLM judge that Turing-RL-trained user simulators improve on human-likeness compared to SFT-Init on both Reddit and Chat.

\paragraph{Are Humans or LLMs Better Judges?} 

\begin{figure}[t]
\centering
\vspace{0.4cm}
\includegraphics[width=\columnwidth]{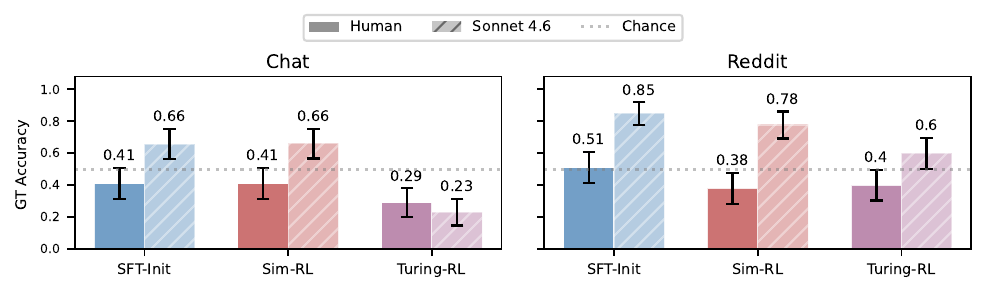}
\caption{\textbf{Comparing human and LLM judge accuracy at identifying the real user's response.} For each of 50 target users per domain, we take the majority vote from ${\sim}6$ human annotators (solid) and from the Sonnet~4.6 Turing judge. \textit{GT Accuracy} is the fraction of targets correctly identified. Both evaluators agree that Turing-RL is the hardest model to distinguish from real users. \textbf{Sonnet~4.6 matches or exceeds human accuracy in most conditions, supporting its use as an automatic evaluation proxy.}}
\label{fig:binary_gt_accuracy}
\end{figure}

\begin{table}[b]
\centering
\small
\setlength{\tabcolsep}{5pt}
\begin{tabular*}{0.75\columnwidth}{@{\extracolsep{\fill}}lcc@{}}
\toprule
Model & Turing & Sim \\
\midrule
SFT-Init & 3.265 & .1791 \\
Sim-RL & 3.645 & .3133 \\
Turing-RL & \textbf{3.857} & \textbf{.3190} \\
\bottomrule
\end{tabular*}
\caption{Out-of-domain transfer to the first 100 MS-TOD users. Higher is better; bold indicates the best result in each column.}
\label{tab:mstod_transfer}
\end{table}

Our human evaluation results are less obviously in favor of the Turing-RL approach. But is it possible that human judges are actually less reliable than the Sonnet 4.6 LLM judge?

We compare the behavior of the LLM judge and human judges by measuring their accuracy in distinguishing human responses from responses generated by different models. Concretely, we binarize the Sonnet~4.6 Turing score for each evaluation thread (score ${<}\,4 \to$ GT, ${>}\,4 \to$ model, ${=}\,4$ excluded) and take the majority vote per target user, mirroring the human forced-choice setup.  Figure~\ref{fig:binary_gt_accuracy} compares the resulting GT accuracy with human majority-vote accuracy across model on 50 target users for each domain. Human annotators recruited under our experimental setup generally underperform Sonnet~4.6, except for judging Turing-RL on Chat, where the results are similar. 

\subsection{Cross-Domain Generalization}
Can models trained to simulate open-ended chat transfer to task-oriented dialogues? We evaluate the three Chat-trained simulators on the first 100 users of MS-TOD, a multi-session task-oriented dialogue benchmark designed to evaluate agents rather than user simulators~\citep{du2026memguide}. As in our Chat evaluation, each simulator conditions on two to four sessions of fixed user history, an inferred persona, and the current context. Table~\ref{tab:mstod_transfer} shows that both RL variants transfer beyond their training domain, with Turing-RL achieving the highest Turing and similarity scores. As MS-TOD provides no established simulator-specific task metrics, we evaluate human-likeness and response similarity rather than goal completion. Appendix Table~\ref{tab:mstod_example} gives a qualitative example.

\section{Ablations and Analysis}
\subsection{User Representations}
User representation is a core aspect of user simulation. As discussed in \S\ref{sec:problem}, we condition our models on both the user behavior data $h$ and the induced persona $\rho = f_{\text{LLM}}(h)$ from an auxiliary LLM. How helpful is the induced persona? We ablate on the user representation in both domains and study three different user representations: $u = h$ (history only), $u = \rho$ (persona only), and $u = (h, \rho)$ (both).

Table~\ref{tab:ablation_combined} shows the results. Turing scores are largely robust to the choice of user representation, while specificity is more domain dependent: Chat remains stable across the three inputs, whereas Reddit favors representations that include history. The fact that persona alone achieves comparable Turing scores, but not comparable similarity scores, suggests that the persona captures stylistic and behavioral patterns sufficient for human-likeness, even when it does not help reproduce the exact content of the ground truth response. We also include the normalized Turing reward curves in [0, 1] during GRPO training for the three input conditions, confirming that training dynamics is not much affected by input types (Figure~\ref{fig:grpo_training_raw_scores}).

To mitigate the bias from a single persona inductor model, we further test whether the user representation results depend on the auxiliary model used to induce the persona $\rho$ in Appendix Table~\ref{tab:persona_inductor_ablation}. We find that larger models do not reliably produce better persona for user representation, and training with $u=(h, \rho)$ does not necessarily improve performance. We leave more systematic investigation of better user representation for future work.
\begin{table}[t]
\centering
\small
\setlength{\tabcolsep}{2.0pt}
\begin{tabular*}{\columnwidth}{@{\extracolsep{\fill}}lcccccc@{}}
\toprule
\multirow{2}{*}{} & \multicolumn{3}{c}{Chat} & \multicolumn{3}{c}{Reddit} \\
\cmidrule(lr){2-4}\cmidrule(l){5-7}
User Representation & Turing & Sim & Specificity & Turing & Sim & Specificity \\
\midrule
$u = (h, \rho)$ (user history + induced persona) & 4.31\subtle{$\pm .08$} & 5.3\subtle{$\pm .8$} & .512\subtle{$\pm .009$} & {3.68}\subtle{$\pm .18$} & 3.0\subtle{$\pm .9$} & .367\subtle{$\pm .020$} \\
$u = h$ (user history only)      & 4.26\subtle{$\pm .08$} & 4.7\subtle{$\pm .8$} & {.497}\subtle{$\pm .010$} & 3.78\subtle{$\pm .18$} & 3.3\subtle{$\pm 1.0$} & .364\subtle{$\pm .019$} \\
 $u = \rho$ (induced persona only)    & {4.23}\subtle{$\pm .08$} & {4.1}\subtle{$\pm .7$} & .509\subtle{$\pm .009$} & 3.84\subtle{$\pm .19$} & {2.7}\subtle{$\pm .7$} & {.329}\subtle{$\pm .018$} \\
\bottomrule
\end{tabular*}
\vspace{-0.3cm}
\caption{Ablation on user representation for Turing-RL. We compare three input conditions: history and persona $u = (h, \rho)$, history only $u = h$, and persona only $u = \rho$.  Values are mean $\pm$ 95\% CI; Turing is on a 1--7 scale, Sim is reported in \%, and Specificity is in $[0,1]$.}
\label{tab:ablation_combined}
\end{table}




\subsection{Qualitative Examples}

\begin{figure}[t]
\centering
\includegraphics[width=1.0\columnwidth]{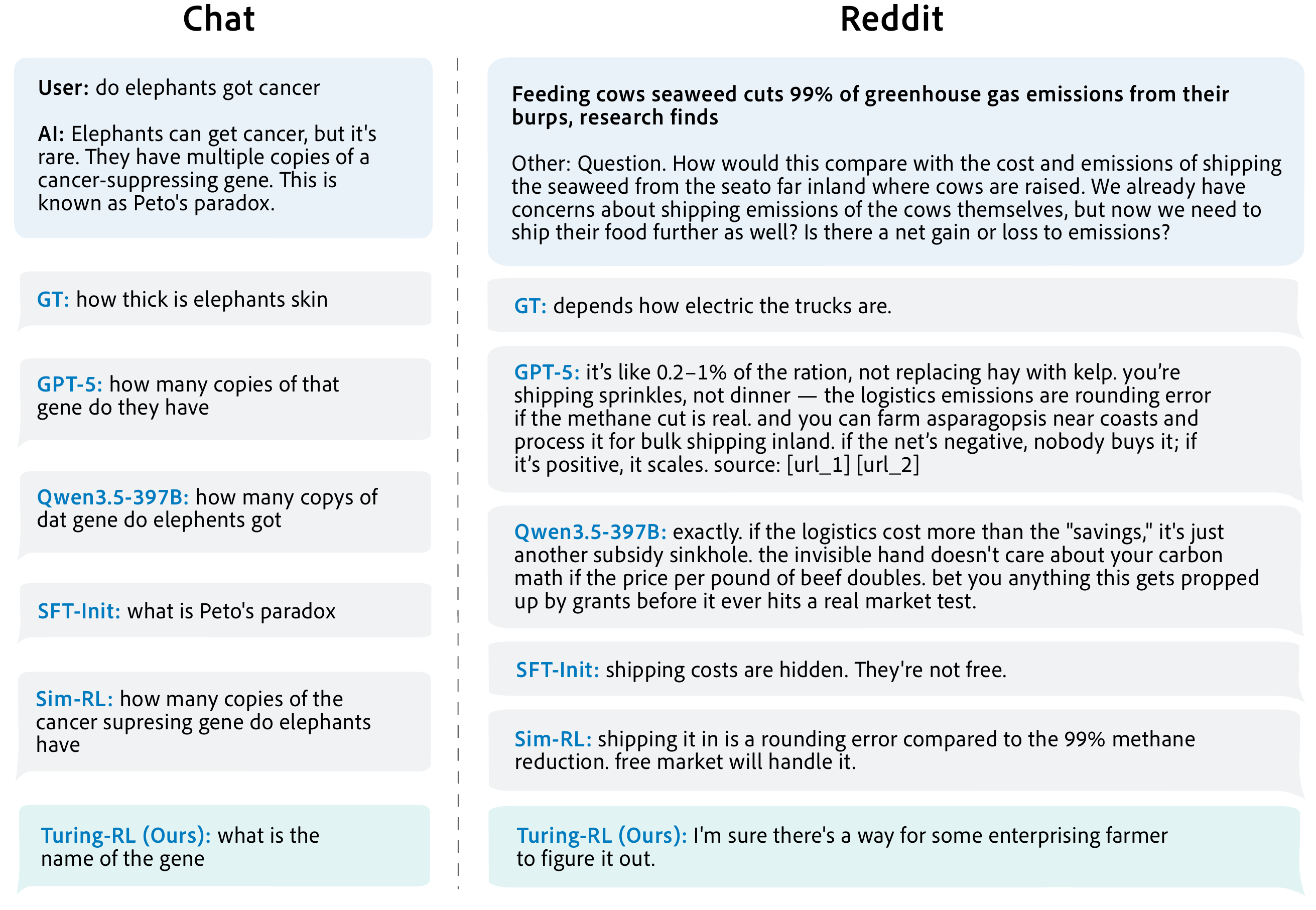}
\vspace{-1em}
\caption{Qualitative comparison of the ground truth, GPT-5, Qwen3.5-397B, SFT-Init, Sim-RL, and Turing-RL on one Chat and one Reddit example.}
\label{fig:qualitative_examples_2}
\vspace{-1em}
\end{figure}

Figure~\ref{fig:qualitative_examples_2} shows representative examples from both domains comparing the ground truth, GPT-5, Qwen3.5-397B, SFT-Init, Sim-RL, and Turing-RL. On Chat, the ground truth is a natural pivot often observed in human-AI conversations. All of the model generations stay anchored to the previous response, and Sim-RL aligns with GPT-5 and Qwen3.5-397B in content. Both SFT-Init and Turing-RL ask plausible follow-ups, although the question raised by SFT-Init is already partially answered in the context. On Reddit, the ground truth is a sharp comment. GPT-5 and Qwen3.5-397B write overly verbose responses in an assistant-like style. SFT-Init responds by paraphrasing a main point in the context. Sim-RL matches GPT-5 in content but is more succinct. Turing-RL provides a plausible human-like reaction. Additional examples including more baselines are shown in Appendix Figure~\ref{fig:qualitative_examples}. We also present a qualitative output from the training-time judge in Appendix~\ref{app:training-judge-example}.

\section{Discussion}
\label{sec:discussion}
A recurring finding across our experiments is that content matching and human-likeness come apart. While similarity reward improves how well a response covers the ground truth content, it does not necessarily make the response harder to distinguish from the real user. The Turing reward, in contrast, improves indistinguishability without lowering content similarity. This suggests that a discriminative signal is better suited than a matching signal for user simulation.

The same capability that makes a simulator useful, however, also poses risks. A model trained to be indistinguishable from a specific person, conditioned on their prior behavior, is by construction well suited to impersonation; it could be used to fabricate convincing messages attributed to real individuals, or to scale fraud and social-engineering attacks. We emphasize that our simulators are trained and evaluated on public or consented research data and are intended for studying interactive systems and human behavior in aggregate, not for reproducing identifiable individuals. While we believe the positive research and application value of user simulation is substantial and outweigh the risks, it should be pursued alongside safeguards such as watermarking generated content and developing AI generation detectors. As with any dual-use technology, we argue that downstream applications should be evaluated in the context of their specific deployment setting and applicable institutional policies to avoid malicious usage. 


There are several directions for future work. First, it would be valuable to study whether user simulators can help LLM assistants become more personalized and better aligned with users' goals and intentions, for example in multi-agent systems \citep{guo2024large, tomavsev2026intelligent} and cognitive architectures \citep{sumers2023cognitive, liu2026cognitive}. Second, while this work focuses on training LLMs to produce \textit{outputs} that are indistinguishable from human-written text, it remains an open question whether the models' reasoning \textit{processes} align with those of humans. One avenue is to compare model-generated reasoning traces with verbalized thought traces from humans \citep{wurgaft2025scaling, kargupta2025cognitive}; another is to explore richer representations for reasoning beyond natural language alone \citep{jha2025modeling, zhang2025code, li2026simulating}. Third, reliable user simulators could enable experiments that elicit group and collective behaviors from populations of simulated agents in open-ended settings, providing a new tool for replicating and extending research in computational social science \citep{lazer2009computational, ziems2024can}.

\section{Related Work}

\paragraph{LLM-based user simulation.}
LMs have been used to evaluate dialogue systems~\citep{davidson2023user, sekulic2024simulating, sim2real, humt}, train conversational agents via self-play~\citep{shah2018bootstrapping, abdulhai2025consistently, utility}, and replicate human subject studies~\citep{aher2023using, park2023generative}. \citet{naous2025flipping} show that assistant-tuned LLMs are structurally misaligned with the user role, while \citet{mehri2025goal} address goal drift in task-oriented simulation through explicit state tracking. These works establish that user simulation benefits from purpose-built training but model generic users rather than specific individuals.
Beyond standard supervised fine-tuning~\citep{wolf2019transfertransfo} for user simulation, \citet{abdulhai2025consistently} use LLM-judged consistency scores as PPO rewards, \citet{wu2026humanlm} optimize for similarity along psychologically grounded dimensions such as stance and emotion, and \citet{gandhi2026dialogue} show that log-probability maximization with latent chain-of-thought outperforms judge-based rewards in generic dialogue prediction. Our discriminative Turing reward takes a different approach: rather than matching content or maximizing likelihood, it trains the model to be indistinguishable from the real user. Concurrent with our work, \citet{OdysSim} build foundation models to simulate human behaviors through task-specific and verbal-feedback-based training \citep{verbal_feedback}, also showing that finetuning a smaller model can beat frontier models on user simulation tasks.

\paragraph{Persona and user representation.}
It is a common practice to use a natural language description of personality, demographics, and traits to represent a character. Persona-conditioned generation was introduced by \citet{zhang2018personalizing} and extended with pretrained models by \citet{wolf2019transfertransfo}, and \citet{jiang2024personallm} show that LLMs can express assigned personality profiles in both self-report and free-form writing. Subsequent work enriches user representations: \citet{hwang2023aligning} show that retrieving relevant past opinions outperforms demographic prompting; \citet{ryan2025synthesizeme} induce synthetic personas from behavioral traces for personalized reward modeling; and \citet{jiang2024indievalue} demonstrate that individual values cannot be approximated by group-level categories. More recently, \citet{xiao2026chameleon} discover that LLM agents assigned distinct profiles still converge into narrow behavioral patterns, providing evidence for the lack of generation diversity in persona prompting. In this work we combine raw history with an induced persona, providing analysis on training models with different user representations.

\paragraph{Human behavior prediction.} Predicting human behaviors, such as people's choices, actions, and utterances, is a cornerstone task in cognitive and social sciences. Classical approaches model human decision-making include symbolic cognitive architectures \citep{newell1994unified, anderson2013adaptive}, the framework of bounded rationality \citep{simon1955behavioral}, and prospect theory that explains heuristics and biases \citep{kahneman2013prospect}. Bayesian cognitive science has become a successful paradigm in predicting human behaviors in controlled experimental settings through building structured models that perform probabilistic inference \citep{griffiths2024bayesian}, which has been applied to studying how people simulate the minds of other people \citep{baker2009action, baker2017rational}, the ability known as Theory of Mind \citep{frith2005theory}.
A complementary, data-driven approach trains neural networks on large datasets of human choices, combining cognitive priors and achieving the strong predictive accuracy \citep{bourgin2019cognitive, peterson2021using}. Nowadays, LLMs have been evaluated as proxies for human participants in surveys, economic games, and social experiments~\citep{argyle2023out, horton2023large, aher2023using, park2024generative}. \citet{binz2025foundation} fine-tune a foundation model on hundreds of psychology experiments to predict trial-level human responses. These approaches have largely focused on constrained, single-turn settings, such as a survey response or an experimental trial, whereas our setting targets open-ended, free-form generations.

\section{Conclusion}

We have proposed training LLM-based user simulators with a discriminative Turing reward, which scores a generated response by how indistinguishable it is from the real user's response, conditioned on the user's prior history. Across two substantially different domains, this signal consistently produces more human-like responses than either log-probability maximization or response-similarity rewards under both LLM- and human-based Turing evaluations, without reducing content alignment with the ground truth. 
Our work suggests a promising recipe for building accurate and generalizable user simulators, with potential applications across a range of downstream settings. 





\section*{Limitations}

We acknowledge the following limitations of our study. First, while we evaluate on two structurally different domains (open-ended chat and forum discussion), there are more extended settings to systematically test the generalization of our Turing reward across other interaction types such as task-oriented dialogue, negotiation, or collaborative problem-solving. Second, our experiments use Qwen3-8B as the base model. While this is sufficient to demonstrate the effectiveness of the Turing reward relative to alternative training signals, it would be valuable to study how our recipe scales to frontier-sized models, where the gap between training signals may narrow or widen in ways we cannot predict from small-model experiments alone. Finally, our discriminative Turing reward relies on a powerful LLM judge (Qwen3.5-397B-A17B) at training time, which introduces both computational cost and a dependence on the judge's own biases. If the judge has systematic blind spots, the trained simulator may learn to exploit those blind spots rather than achieve genuine human-likeness. Our human evaluation partially addresses this concern by showing that improvements transfer to real human judges, but a more thorough analysis of judge–human disagreement patterns would be valuable for future work.


\section*{Acknowledgments}
This study was supported by MIT-IBM Watson AI Lab. We thank the Modal credit grant for academics program (https://modal.com/academics) for providing some of the cloud computing GPU resources.
\bibliography{references}
\bibliographystyle{colm2026_conference}

\clearpage

\appendix

\section{Dataset Details}
\label{app:dataset-stats}

Each training example is a tuple $(u, x, y)$ where the user representation $u = (h, \rho)$ is fixed per user and disjoint from both the current context $x$ and the ground truth response $y_{\star}$. The behavioral history $h$ is selected once per user from a deterministic seed; the persona $\rho$ is induced from $h$ alone. This separation guarantees that neither $h$ nor $\rho$ leaks information about the target.

\subsection{Dataset Preprocessing}

We show the dataset statistics after preprocessing in Table~\ref{tab:preprocessing_stats} below.

\begin{table}[h]
\centering
\small
\begin{tabular}{lrr}
\toprule
 & PRISM & ConvoKit \\
\midrule
Users & 1,288 & 1,282 \\
History & 2--4 conv. & 2--6 threads \\
SFT train rows & 3,272 & 4,264 \\
GRPO train rows & 4,174 & 4,737 \\
GRPO val rows & 705 & 1,661 \\
Test rows & 880 & 267 \\
\bottomrule
\end{tabular}
\caption{Dataset statistics after preprocessing. History counts are sampled per user. PRISM and ConvoKit use disjoint SFT, GRPO, and held-out test users; the held-out ConvoKit test set comes from \texttt{r/tifu} and \texttt{r/worldnews}.}
\label{tab:preprocessing_stats}
\end{table}

\paragraph{(Chat) PRISM.}
This dataset is distributed under the cc license. We retain PRISM users with at least six conversations, yielding 1,288 qualified users. We randomly hold out 128 qualified users with a fixed seed, yielding 128 held-out test users and 880 test examples. The remaining 1,160 users are split with the same seed into disjoint GRPO and SFT sets using a 60/40 user ratio: 696 users for GRPO and 464 users for SFT. For each user, 2--4 conversations are reserved as $h$; the rest serve as targets. Because PRISM dialogues are linear, each user turn in a target conversation yields one example: $x$ is all preceding turns, and $y$ is the user's utterance at that turn.

\paragraph{(Reddit) ConvoKit.}
This dataset is distributed under the MIT License. We reconstruct Reddit threads from comment identifiers and parent pointers.
We reserve \texttt{r/tifu} and \texttt{r/worldnews} as held-out test subreddits and remove their users from both SFT and GRPO training.
We use the remaining subreddits for SFT and GRPO training. These include \texttt{r/AmItheAsshole}, \texttt{r/AskMen}, \texttt{r/AskWomen}, \texttt{r/business}, \texttt{r/changemyview}, \texttt{r/Economics}, \texttt{r/Frugal}, \texttt{r/MaliciousCompliance}, \texttt{r/news}, \texttt{r/relationship\_advice}, \texttt{r/relationships}, and \texttt{r/TrueReddit}.

Users are deterministically split into disjoint SFT and GRPO sets at a 40/60 ratio.
For each user, 2--6 threads are reserved as $h$, and users with insufficient threads are discarded. Each target thread yields one example: $y_{\star}$ is the user's final comment, and $x$ is the original post plus the ancestor reply chain leading to that comment.




\subsection{Persona Induction}

Our default user representation pairs the user's behavioral history with an induced persona. To induce a persona, we prompt GPT-5.4 nano (temperature 0.2, max output length 2{,}048 tokens) to summarize each user's history block $h$ into a fixed persona $\rho$ with five fields: values, verbal quirks, expression style, length prior, and background. The same $\rho$ is reused for every example belonging to that user. Figure~\ref{fig:persona_induction} shows the induction prompt.

\begin{figure*}[t]
\centering

\begin{promptbox}[System Message (Persona Induction)]
You write compressed first-person persona notes as if the target user wrote them about their own habits.
Sound like the user, not like an analyst, therapist, teacher, or policy brief.
Output only the requested strict JSON object and nothing else.
\end{promptbox}

\begin{promptbox}[Persona Induction Prompt (Part 1/2)]
<target_speaker>[HUMAN]</target_speaker>
<selected_history>
(reserved history block (*@$h$@*), with [HUMAN]/[OTHER] speaker labels and bold-delimited target-user responses)
</selected_history>

<task>
Write a compact persona for [HUMAN] in [HUMAN]'s own voice.

Rules:
- Write each field in first person, as if [HUMAN] is describing their own habits.
- Keep it casual and compressed, not polished or formal.
- Use only [HUMAN]'s own messages as evidence.
- Capture only stable traits that are useful for predicting future messages across contexts.
- Prefer concrete wording habits over abstract summaries.
- Do not sound like an analyst describing a person from the outside.
- Do not use therapist/assistant/policy-brief language.
- Avoid abstract labels such as fairness, accountability, autonomy, respect, boundaries, power dynamics, empathy, nuance, skepticism, advocacy, procedural correctness, or similar summary words unless those exact words are frequent in the history.
- Do not infer demographics or sensitive attributes.
- Do not infer motives, worldview, morals, personality traits, or values unless they are explicitly and repeatedly stated.
- Do not include one-off topical stances, transient emotions, or local interaction goals.
\end{promptbox}

\vspace{-6pt}

\caption{Persona induction prompt (part 1 of 2). The inducer receives the user's reserved history $h$ formatted with speaker labels and bold-delimited target-user responses, and outputs a structured JSON persona $\rho$.}
\label{fig:persona_induction}
\end{figure*}

\begin{figure*}[t]
\centering

\begin{promptbox}[Persona Induction Prompt (Part 2/2)]
Rules (continued):
- If evidence is weak, write `unknown`.
- Keep each field concise.
- Write each field as short fragments or very short sentences, not polished mini-paragraphs.
- Do not quote or reproduce complete sentences from the selected history.
- Do not quote or reuse exact words or phrases from the selected history, including verdict labels, slang, catchphrases, or short text fragments.
- Do not include literal examples from [HUMAN]'s messages, even single-word examples.
- Describe lexical habits generically rather than by repeating exact tokens.
- Non-text segments such as punctuation styles may be described, but do not quote surrounding text.
- Do not include parenthetical example lists, `e.g.` scaffolding, or full sample sentences from [HUMAN]'s messages.

Field descriptions:
- values: only concrete recurring preferences, irritations, or things I keep siding with; avoid abstract moral language unless it is explicit in the history
- verbal_quirks: my observable surface-form habits only, such as how I use interjections, punctuation, grammar looseness, discourse markers, formatting quirks, or short repeated lexical patterns. Do not include exact verdict labels or quoted wording from the history.
- expression_style: how I usually come across at the sentence level; keep this concrete and plain, not psychological, evaluative, or therapist-like
- length_prior: my usual reply length in plain language, mainly sentence count and rough brevity
- background: concrete background cues not already captured above, especially repeated personal experiences, self-disclosed roles, responsibilities, constraints, routines, domain familiarity, sources of information, or recurring points of reference that likely shape future responses; do not restate general values, tone, or reasoning style; use `unknown` if nothing strong remains

Output only this JSON object shape:
{"values": "...", "verbal_quirks": "...", "expression_style": "...", "length_prior": "...", "background": "..."}
</task>
\end{promptbox}

\begin{promptbox}[Model Output]
{"values": "...", "verbal_quirks": "...", "expression_style": "...", "length_prior": "...", "background": "..."}
\end{promptbox}

\vspace{-6pt}

\caption{Persona induction prompt (part 2 of 2). Continuation of rules and field descriptions, followed by the expected model output schema.}
\label{fig:persona_induction_2}
\end{figure*}

\section{Generation Details}\label{sec:sampling_params}
\subsection{Sampling Parameters}
The sampling parameters used for user simulation are listed in Table~\ref{tab:heldout_generation_params}.

\begin{table}[h]
\centering
\small
\setlength{\tabcolsep}{3.2pt}
\begin{adjustbox}{max width=\linewidth}
\begin{tabular}{@{}llcccccc@{}}
\toprule
Model & Backend & Max tokens & Temp. & Top-$p$ & Top-$k$ & Pres. pen. & Rep. pen. \\
\midrule
Qwen3-8B & vLLM & 2048 & 0.4 / 0.6 & 1.0 & -1 & 0.5 & 1.0 \\
GPT-5 & API & 1024 & -- & -- & -- & -- & -- \\
Qwen3.5-397B-A17B & vLLM & 1024 & 0.6 & 1.0 & -1 & 0.5 & -- \\
\bottomrule
\end{tabular}
\end{adjustbox}
\caption{Held-out generation sampling parameters. For Qwen3-8B, Reddit generations use temperature 0.4 and Chat generations use temperature 0.6. All Qwen3-8B rows use one generation per target.}
\label{tab:heldout_generation_params}
\end{table}

\subsection{Prompt Formatting}
Each example is rendered as a two-message chat sequence: a system message containing the simulation instruction, the task description, and (when applicable) the induced persona $\rho$ (Figure~\ref{fig:system_prompt}); and a user message containing the behavioral history $h$ followed by the current context $x$. The model generates a chain-of-thought trace $z$ enclosed in \texttt{<reasoning>} tags, then the response $y$ prefixed by \texttt{[HUMAN]:}. Figures~\ref{fig:prism_format} and~\ref{fig:convokit_format} show the user message layout for each dataset.
\begin{figure*}[t]
\centering

\begin{promptbox}[System Message (User Simulation)]
You are simulating a human labeled by [HUMAN]. Your job is to predict what [HUMAN] says as if you are that person.

[PERSONA]
<persona>
    <values>...</values>
    <verbal_quirks>...</verbal_quirks>
    <expression_style>...</expression_style>
    <length_prior>...</length_prior>
    <background>...</background>
</persona>

[TASK]
Your task is to predict [HUMAN]'s next message, matching [HUMAN]'s writing intentions, style, vocabulary, and tone.
Use all provided information about [HUMAN] and the current context to predict the next message.
Target your message to [OTHER] or [OTHER - OP], not to people who are described in the context but are not participants in the thread or conversation. Do not answer as an assistant, analyst, or narrator.

Before outputting the final answer, briefly reason about what [HUMAN] would say and think step by step.
The generated response should naturally follow from your reasoning.
Format your output exactly like this:
<reasoning>...</reasoning>
[HUMAN]: your response
Write any reasoning before `[HUMAN]:` enclosed by the reasoning tags, and write only the final response after `[HUMAN]:`.
Do not include reasoning after `[HUMAN]:`.
\end{promptbox}

\vspace{-6pt}

\caption{System message shared across all prompt configurations. The \texttt{[PERSONA]} block is included only in persona-conditioned settings; the \texttt{[TASK]} block specifies the prediction objective and output format.}
\label{fig:system_prompt}
\end{figure*}

\begin{figure*}[t]
\centering

\begin{promptbox}[User Simulation Prompt (PRISM)]
[USER HISTORY]
Below are previous conversations involving [HUMAN]. Use them to infer [HUMAN]'s style, values, preferences, and likely responses.

<Conversation 1>
[MESSAGES]
[HUMAN]: (user turn)
[OTHER]: (assistant turn)
[HUMAN]: (user turn)
[OTHER]: (assistant turn)
</Conversation 1>

[CURRENT CONTEXT]
[MESSAGES SO FAR]
[HUMAN]: (turn 0)
[OTHER]: (turn 0)
[HUMAN]: (turn 1)
[OTHER]: (turn 1)
\end{promptbox}

\begin{promptbox}[Model Output]
<reasoning>...</reasoning>
[HUMAN]: (*@$\hat{y}$@*)
\end{promptbox}

\vspace{-6pt}

\caption{User message layout for PRISM (multi-turn dialogue). History conversations are wrapped in numbered \texttt{<Conversation>} tags; the current context lists turns up to the prediction point.}
\label{fig:prism_format}
\end{figure*}

\begin{figure*}[t]
\centering

\begin{promptbox}[User Simulation Prompt (ConvoKit)]
[USER HISTORY]
Below are [HUMAN]'s previous messages.

<Post 1>
[OTHER - OP]
(original post text)

[OTHER]: (reply)
[HUMAN]: (reply)
[OTHER]: (reply)
</Post 1>

[CURRENT CONTEXT]
[OTHER - OP]
(original post text)

[OTHER]: (ancestor reply)
[HUMAN]: (earlier comment)
[OTHER]: (ancestor reply)
[HUMAN]:
\end{promptbox}

\begin{promptbox}[Model Output]
<reasoning>...</reasoning>
[HUMAN]: (*@$\hat{y}$@*)
\end{promptbox}

\vspace{-6pt}

\caption{Prompt layout for ConvoKit (Reddit forum). History threads are wrapped in numbered \texttt{<Post>} tags; the original poster is marked \texttt{[OTHER - OP]}. The current context shows the OP and ancestor reply chain up to the prediction point.}
\label{fig:convokit_format}
\end{figure*}

\section{Training Details}
\label{app:training_details}
\subsection{SFT Training Details}
\label{app:sft_training_details}
\label{app:thinking-trace-prompt}
\label{app:hyperparams}

The user splits reserved for SFT training are listed in Table~\ref{tab:preprocessing_stats}. Each SFT training example pairs a ground truth user response with a generated reasoning trace \citep{lu2026llmagentssimulatemultiturn}. To construct each thinking trace, we run Qwen3-8B with thinking mode disabled and provide the user's reserved history $h$, induced persona $\rho$, current context $x$, and ground truth response $y_{\star}$. The prompt asks the model to explain what could have led \texttt{[HUMAN]} to write the response, focusing on the local target, intent, stance, style, and plausible length, while explicitly prohibiting copying or closely paraphrasing the ground truth response. We additionally run a ground truth leakage check and regenerate any trace flagged for exposing the wording of $y_{\star}$ until the check passes. The exact thinking-trace generation prompt is shown in Figure~\ref{fig:thinking_trace_prompt}. The resulting trace is paired with the ground truth as \texttt{<reasoning>}$t$\texttt{</reasoning> [HUMAN]:}~$y_{\star}$. The loss is applied only to this assistant completion target, while prompt, history, persona, and contexts are masked out. SFT training is implemented with Transformers, TRL SFTTrainer, and PEFT LoRA/QLoRA adapters \citep{transformers, TRL, peft}. Table~\ref{tab:sft_hyperparams} lists the hyperparameters for the SFT warm-start stage.

\begin{figure*}[t]
\centering
\begin{promptbox}[System Message (Thinking Trace Generation)]
You generate hidden reasoning traces for user-simulation SFT data.
Given [HUMAN]'s history, [HUMAN]'s persona, the current
conversation context, and [HUMAN]'s actual next response,
explain the reasoning process that could lead [HUMAN]
to write that response.
\end{promptbox}

\vspace{15pt}

\begin{promptbox}[Thinking Trace Generation Prompt]
<|Target User Evidence|>
(reserved history block (*@$h$@*))
<|End Target User Evidence|>
<|Persona|>
(induced persona (*@$\rho$@*))
<|End Persona|>
<|Interaction Context|>
(current context (*@$x$@*))
<|End Interaction Context|>
<|Given Response|>
(ground-truth response (*@$y$@*))
<|End Given Response|>
Generate the hidden reasoning trace that could lead
[HUMAN] to write the given response.
The trace should reason carefully about:
- what [OTHER] or [OTHER - OP] most recently said
- what exact point [HUMAN]'s response should target
- [HUMAN]'s likely stance, emotion, goal, and style
- plausible response length and level of detail
Bring in supporting evidence from [HUMAN]'s history messages or persona.
Do not reproduce the given response. Do not copy, quote, closely paraphrase, or reveal its final wording. The trace may describe the response's intent, stance, target, and style, but must not restate the response itself.
Output only the reasoning trace text, with no XML tags or headings.
\end{promptbox}
\caption{
Thinking trace elicitation prompt. A teacher model receives the user's history $h$, persona $\rho$,
current context $x$, and ground-truth response $y$, and generates
a reasoning trace explaining what could lead the user to write
that response without reproducing it.
}
\label{fig:thinking_trace_prompt}
\end{figure*}

\begin{table}[h]
\centering
\small
\begin{tabular}{ll}
\toprule
\textbf{Parameter} & \textbf{Value} \\
\midrule
Base model & Qwen3-8B (thinking disabled)\\
Backend & TRL SFTTrainer + Transformers/PEFT LoRA\\
LoRA rank / alpha & 64 / 128 \\
LoRA dropout & 0.05 \\
Learning rate & $2 \times 10^{-4}$ \\
LR scheduler & cosine \\
Max sequence length & 8192 tokens \\
Per-device batch size & 1 \\
Gradient accumulation steps & 16 \\
Number of GPUs & 8 \\
Effective batch size & 128 \\
Total epochs & 3 \\
Precision & bfloat16 \\
\bottomrule
\end{tabular}
\caption{SFT training hyperparameters.}
\label{tab:sft_hyperparams}
\end{table}

\subsection{GRPO Training Details}
\label{app:grpo_training_details}

Table~\ref{tab:hyperparams} lists the hyperparameters shared across GRPO variants. All GRPO runs are initialized from the SFT checkpoints and trained on a GRPO user split disjoint from the SFT data (see Table~\ref{tab:preprocessing_stats}). We sample four candidate responses per example, compute group-relative advantages within each candidate group, and optimize the resulting policy objective with LoRA adapters. All GRPO trainings are veRL-based \citep{verl}. Across SFT and GRPO, we train on B200/B300 machines for about 1680 GPU hours.

We optimize the GRPO objective of \citet{shao2024deepseekmath}. For each prompt $q$, GRPO samples a group of $G$ outputs $\{o_1,\dots,o_G\}$ from the old policy $\pi_{\theta_{\mathrm{old}}}$ and updates the policy $\pi_\theta$ by maximizing
\begin{equation*}
\begin{aligned}
&\mathcal{J}_{\mathrm{GRPO}}(\theta) = \mathbb{E}_{q \sim P(Q),\ \{o_i\}_{i=1}^{G} \sim \pi_{\theta_{\mathrm{old}}}(O \mid q)} \\
&\resizebox{\linewidth}{!}{$\Bigg[ \displaystyle \frac{1}{G}\sum_{i=1}^{G}\frac{1}{|o_i|}\sum_{t=1}^{|o_i|}\bigg\{ \min\!\Big[ \frac{\pi_\theta(o_{i,t}\mid q,o_{i,<t})}{\pi_{\theta_{\mathrm{old}}}(o_{i,t}\mid q,o_{i,<t})}\,\hat{A}_{i,t},\ \mathrm{clip}\!\Big(\frac{\pi_\theta(o_{i,t}\mid q,o_{i,<t})}{\pi_{\theta_{\mathrm{old}}}(o_{i,t}\mid q,o_{i,<t})},\, 1-\varepsilon,\, 1+\varepsilon\Big)\,\hat{A}_{i,t} \Big] - \beta\, \mathbb{D}_{\mathrm{KL}}\!\big[\pi_\theta \,\|\, \pi_{\mathrm{ref}}\big] \bigg\}\Bigg] ,$}
\end{aligned}
\end{equation*}
where $\varepsilon$ is the clipping range and $\beta$ the KL coefficient. More specifically:
\begin{itemize}
\item $q \sim P(Q)$: a training prompt drawn from the GRPO user split. $P(Q)$ is the empirical distribution over these prompts.
\item $\{o_i\}_{i=1}^{G}$: the $G=4$ candidate generations sampled per prompt. Each $o_i = (z_i, \hat{y}_i)$ is a reasoning trace $z_i$ followed by the response $\hat{y}_i$. $o_{i,t}$ is its $t$-th token.
\item $\pi_\theta$: the simulator policy being optimized.
\item $\pi_{\theta_{\mathrm{old}}}$: the behavior (sampling) policy that produced the group $\{o_i\}$, namely the policy from immediately before the current update; the per-token ratio $\pi_\theta/\pi_{\theta_{\mathrm{old}}}$ is the importance weight. Because we take a single PPO epoch per batch (Table~\ref{tab:hyperparams}), $\pi_{\theta_{\mathrm{old}}}=\pi_\theta$ at the first inner step.
\item $\pi_{\mathrm{ref}}$: the SFT checkpoint (Appendix~\ref{app:sft_training_details}) that initializes RL. The penalty $\mathbb{D}_{\mathrm{KL}}\!\big[\pi_\theta \,\|\, \pi_{\mathrm{ref}}\big]$ coefficient is $\beta=1\times10^{-3}$ (Table~\ref{tab:hyperparams}).
\item $\varepsilon$: the PPO clipping range, set to $0.2$ (Table~\ref{tab:hyperparams}).
\item $\hat{A}_{i,t}$: the advantage of token $t$ in output $o_i$. $\hat{A}_{i,t}=\big(r_i-\mathrm{mean}(\mathbf{r})\big)/\mathrm{std}(\mathbf{r})$, where $r_i$ is individual response reward and $\mathbf{r}=\{r_1,\dots,r_G\}$ are the group rewards. The reasoning tokens $z_i$ receive the same advantage.
\end{itemize}

\begin{table}[h]
\centering
\small
\begin{tabular}{ll}
\toprule
\textbf{Parameter} & \textbf{Value} \\
\midrule
\multicolumn{2}{l}{\emph{Shared across all GRPO variants}} \\
Base model & Qwen3-8B (thinking disabled)\\
Backend & veRL \\
Temperature & 0.6 \\
Top \textit{p} & 1.0 \\
Top \textit{k} & -1 \\
Initialization & SFT checkpoint \\
LoRA rank / alpha & 64 / 32 \\
Learning rate & $1 \times 10^{-5}$ \\
KL coefficient ($\beta$) & $1 \times 10^{-3}$ \\
Generations per example ($G$) & 4 \\
Train batch size & 128 \\
PPO epochs per batch & 1 \\
Total epochs & 3 \\
Max response length & 1024 tokens \\
PPO clip ratio & 0.2 \\
Loss Aggregation & token mean \\

\bottomrule
\end{tabular}
\caption{GRPO training hyperparameters shared across all reward variants.}
\label{tab:hyperparams}
\end{table}

\subsection{Length Penalty}
\label{app:length_penalty}

Our GRPO training incorporates a length penalty for responses that fall outside a tolerance band around the ground truth length.
Let $\ell$ be the generated response length, $\ell_{\mathrm{gt}}$ the ground truth response length, and $r = \ell / \ell_{\mathrm{gt}}$ the length ratio. We define an acceptable length-ratio range $[r_{\min}, r_{\max}]$ within which no penalty is applied. Outside this range, the penalty increases linearly with the relative violation, assigning a larger weight to short responses:
\begin{align}
p &= \min\{\lambda_{\mathrm{short}} v_{\mathrm{short}}
        + \lambda_{\mathrm{long}} v_{\mathrm{long}},\; p_{\mathrm{cap}}\},
\end{align}

where $v_{\mathrm{short}} = \max((r_{\min} - r) / r_{\min},\; 0)$ and $v_{\mathrm{long}} = \max((r - r_{\max}) / r_{\max},\; 0)$. Table~\ref{tab:reward-shaping} lists the dataset-specific parameter values. We set these values based on responses generated by the SFT model.

\begin{table}[h]
\centering
\small
\begin{tabular}{lcc}
\toprule
\textbf{Parameter} & \textbf{Reddit} & \textbf{Chat} \\
\midrule
$[r_{\min}, r_{\max}]$ & $[0.8, 1.1]$ & $[0.6, 1.4]$ \\
$\lambda_{\mathrm{short}}$ & 0.45 & 0.40 \\
$\lambda_{\mathrm{long}}$ & 0.15 & 0.20 \\
Penalty cap $p_{\mathrm{cap}}$ & 0.25 & 0.40 \\
\bottomrule
\end{tabular}
\caption{Length penalty parameters for each domain.}
\label{tab:reward-shaping}
\end{table}

\section{Human Evaluation Details}
\label{app:human-eval}

\begin{figure*}[t]
\centering
\includegraphics[width=0.75\textwidth]{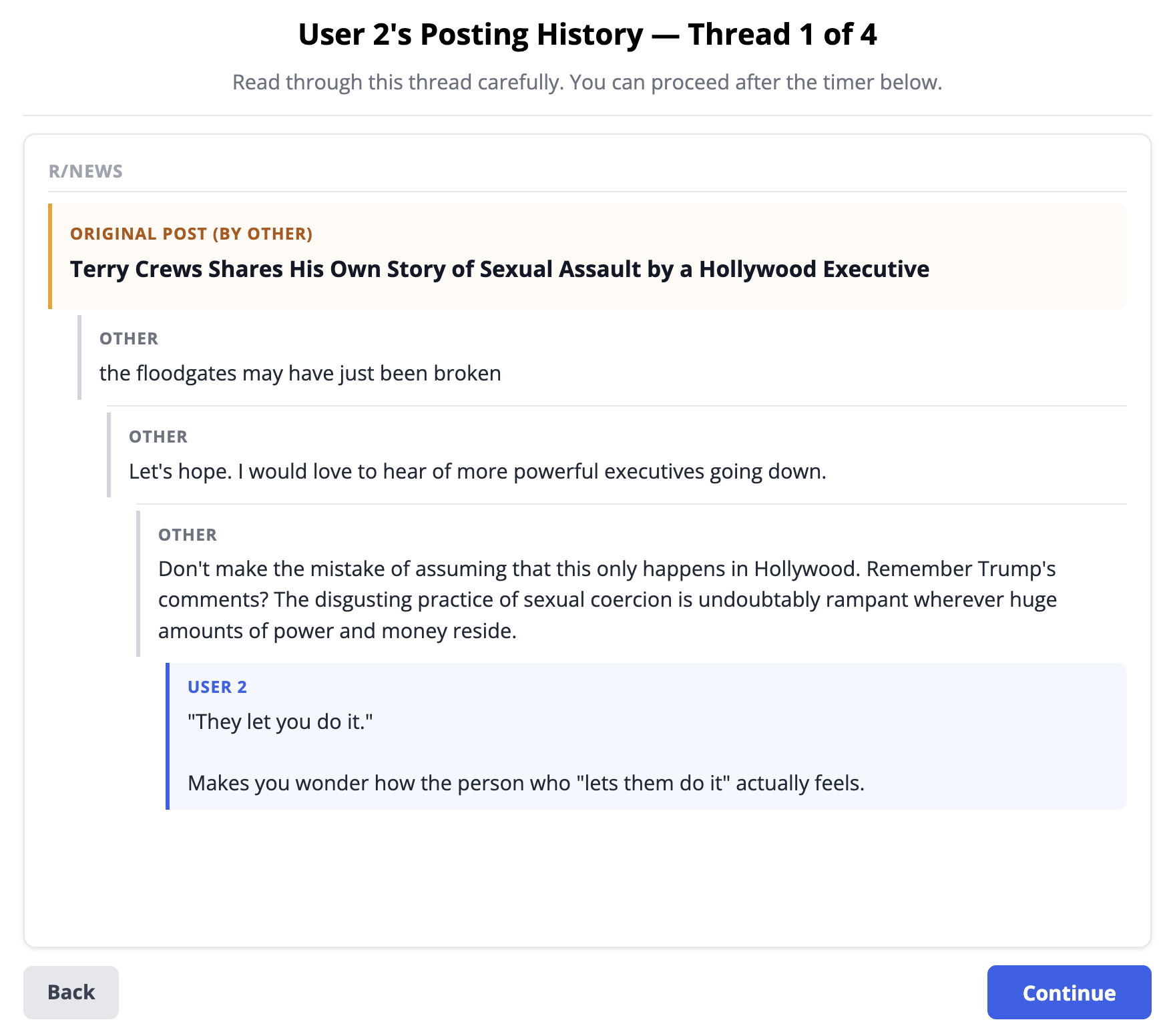}
\includegraphics[width=\textwidth]{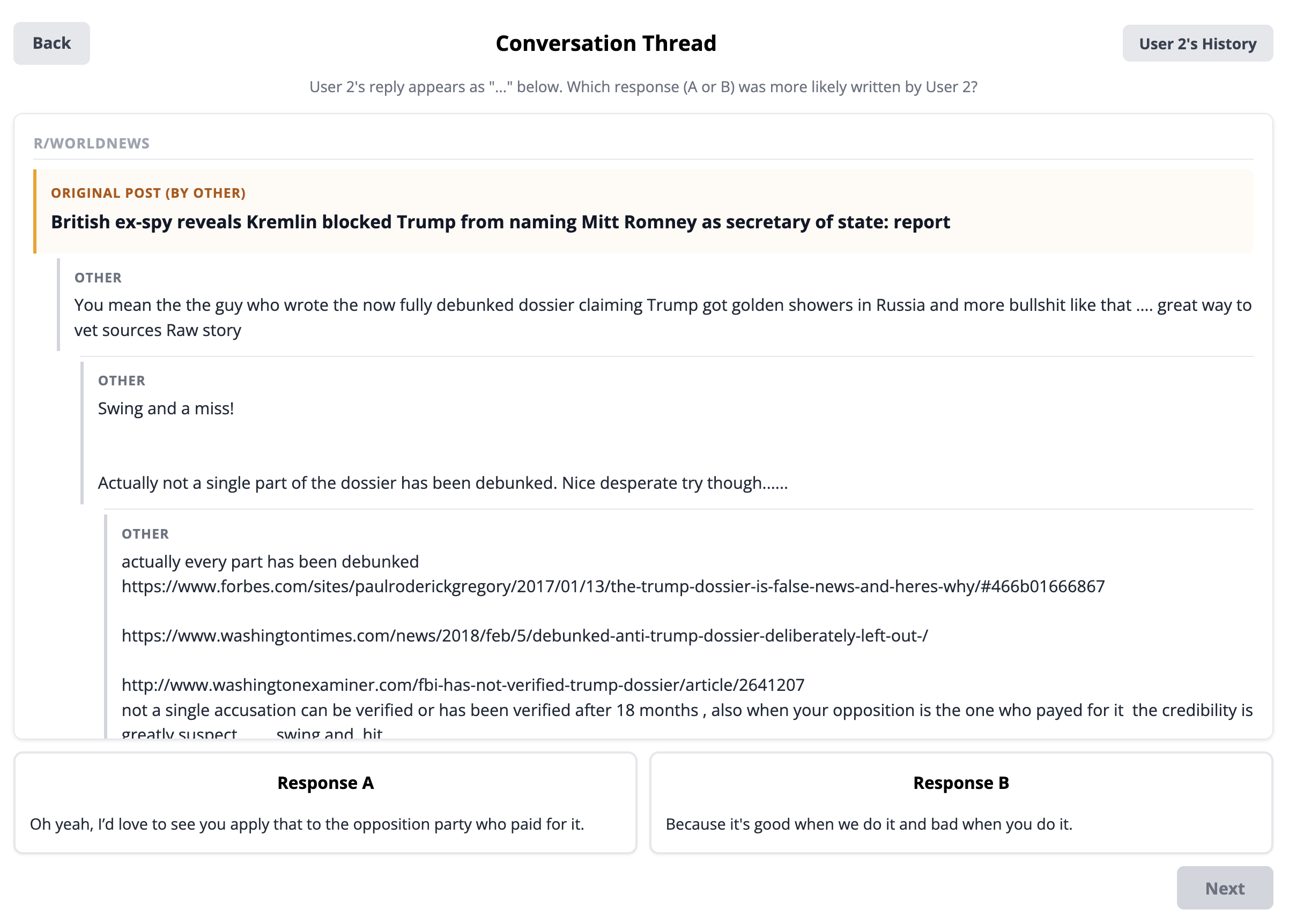}
\caption{User interface used for human annotation in the Reddit domain.}
\label{fig:ui_reddit}
\end{figure*}

\begin{figure*}[t]
\centering
\includegraphics[width=0.75\textwidth]{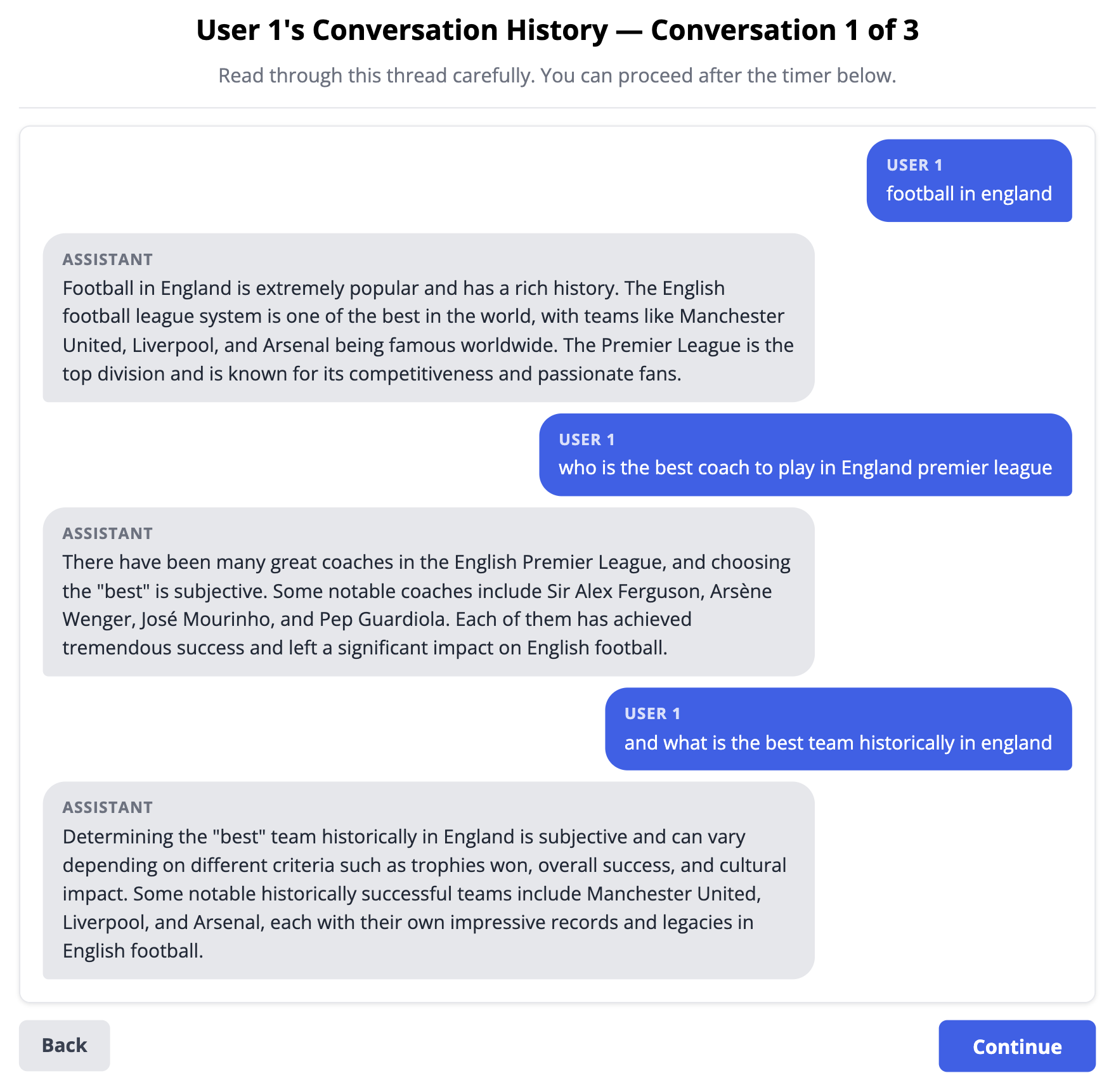}
\includegraphics[width=\textwidth]{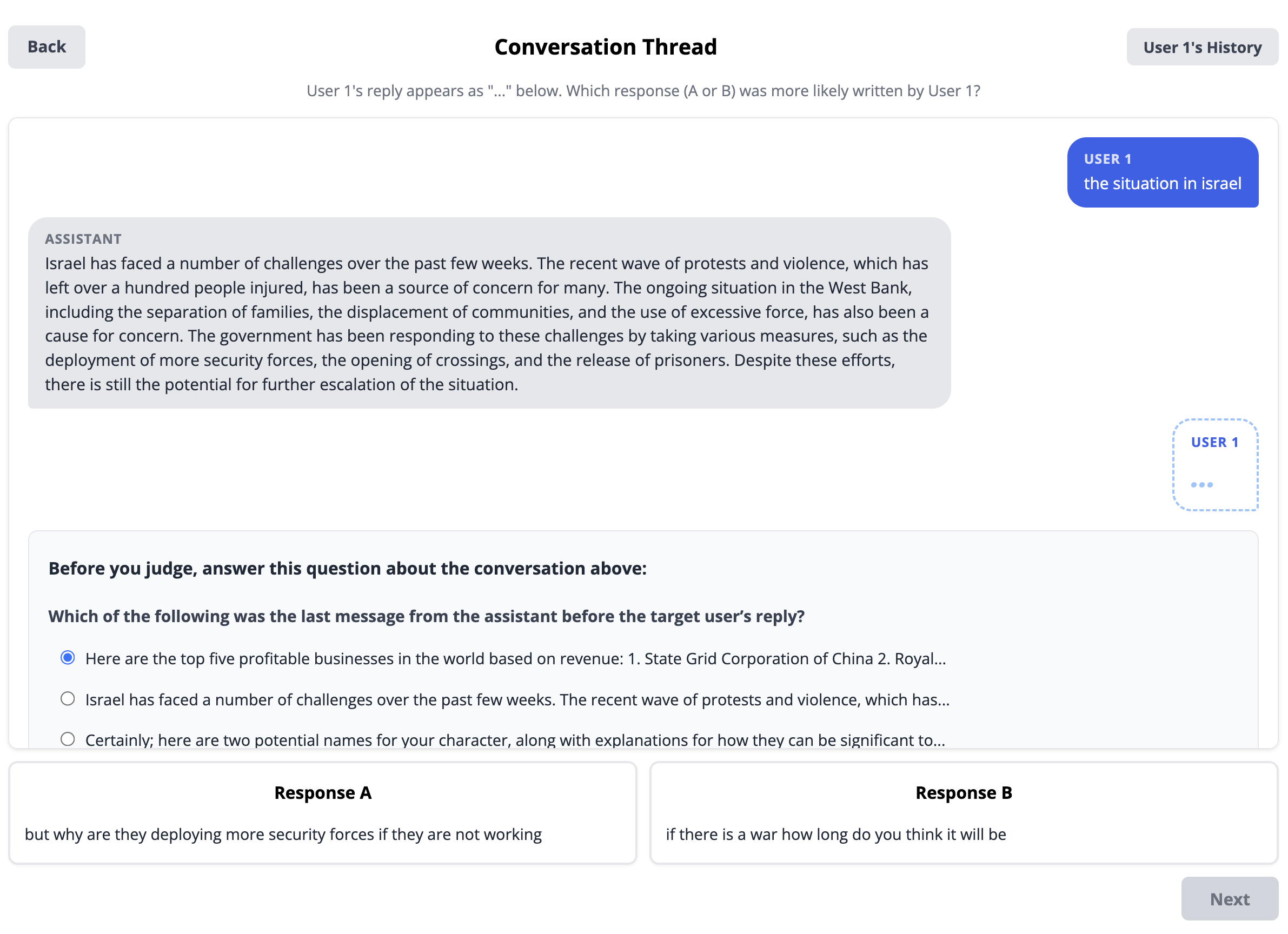}
\caption{User interface used for human annotation in the Chat domain.}
\label{fig:ui_chat}
\end{figure*}

We conduct a forced-choice Turing test on Prolific. Each participant views a target user's history and then judges 10 response pairs: one ground truth response and one model-generated response, presented in randomized order. The participant selects which response they believe was written by the real user. Figures~\ref{fig:ui_reddit} and~\ref{fig:ui_chat} show the annotation interfaces for the Reddit and Chat domains, respectively. Each participant is paid \$6 and the human evaluation costs approximately \$2880 in total (including Prolific service fee and taxes). 

\subsection{Design}
We evaluate three model variants: SFT-Init, Sim-RL, and Turing-RL, on two datasets: ConvoKit (Reddit) and PRISM (Chat). For each dataset, we select 100 target users and split them into 10 groups of 10. Each participant is assigned to one dataset--model--group cell and evaluates the 10 targets in that group. We recruit 6 annotators per cell, yielding 60 participants and 600 binary judgments for each dataset--model pair.

\subsection{Comprehension Filtering}
Each trial includes comprehension checks that test whether the participant read the target user's history, such as identifying the user's most-discussed topic or most recent message. We compute each participant's comprehension accuracy across all check questions and exclude participants scoring below 75\%. 

\subsection{Balancing}
After excluding incomplete submissions, defined as submissions with fewer than 10 judgments, and submissions from participants who fail the comprehension filter, we retain the first 6 valid completions for each dataset, model, and user group, ordered by initialization time. This yields exactly 60 participants and 600 votes for each dataset and model. 

\subsection{Metric}
We report the model win rate (WR): the fraction of judgments in which the annotator chose the model-generated response over ground truth. WR is computed per target user (averaging over the ${\sim}6$ annotators for that target) and then averaged across the 100 targets. The 95\% confidence interval is $\text{mean} \pm 1.96 \times \text{SEM}$, where SEM is the standard error of the 100 per-target win rates.

\subsection{Testing for Statistical Significance}
\label{app:human-eval-significance}
We test pairwise model differences at the heldout-target level. For each dataset and target $i$, we compute the selected-response win rate for each model and form paired differences $d_i = \text{WR}_{i,\text{Turing-RL}} - \text{WR}_{i,\text{comparison}}$ over the 100 targets per dataset. Table~\ref{tab:human_eval_significance} reports the mean paired difference. Bootstrap intervals resample the 100 targets with replacement $10^6$ times and recompute the mean difference. The table reports the 2.5--97.5 percentile interval. 

We then perform paired permutation test. Under the null that the two models perform equally well, each target-level difference $d_i$ is just as likely to favor either model. For each target with nonzero $d_i$, we reassign the difference to be $\pm d_i$. Thus, for $m$ nonzero $d_i$s, there are $2^m$ possible assignments. For each such assignment, we recompute the mean paired difference. The two-sided p-value is the fraction of assignments whose mean difference has absolute value at least as large as the observed mean difference.

\begin{table}[t]
\centering
\small
\setlength{\tabcolsep}{3.5pt}
\begin{adjustbox}{max width=\linewidth}
\begin{tabular}{@{}llcc@{}}
\toprule
Dataset & Comparison & Mean diff & Paired permutation $p$ \\
\midrule
Chat & Turing-RL - SFT-Init & $+.073$\subtle{$\pm .068$} & $0.044$ \\
Chat & Turing-RL - Sim-RL & $+.072$\subtle{$\pm .058$} & $0.022$ \\
Reddit & Turing-RL - SFT-Init & $+.088$\subtle{$\pm .063$} & $0.0095$ \\
Reddit & Turing-RL - Sim-RL & $-.018$\subtle{$\pm .063$} & $0.61$ \\
\bottomrule
\end{tabular}
\end{adjustbox}
\caption{Paired target-level significance tests for Table~\ref{tab:human_eval_winrate}. Mean differences are Turing-RL win rate minus the comparison model's win rate; $\pm$ values are 95\% bootstrap CI.}
\label{tab:human_eval_significance}
\end{table}

\subsection{Word Counts}
We summarize the amount of text shown for each target in Table~\ref{tab:human_eval_word_counts}, counting words in the displayed user history and target context, excluding response options and interface text.

\begin{table}[t]
\centering
\small
\setlength{\tabcolsep}{4pt}
\begin{adjustbox}{max width=\linewidth}
\begin{tabular}{@{}lccc@{}}
\toprule
Dataset & Mean Words/Question & History Mean & Context Mean \\
\midrule
Chat & 1303.3\subtle{$\pm 140.4$} & 1106.6\subtle{$\pm 129.8$} & 196.7\subtle{$\pm 40.2$} \\
Reddit & 1676.6\subtle{$\pm 306.4$} & 1328.1\subtle{$\pm 282.5$} & 348.5\subtle{$\pm 63.1$} \\
\bottomrule
\end{tabular}
\end{adjustbox}
\caption{Word counts for the displayed human-evaluation stimuli. Values show mean words per target with 95\% CI.}
\label{tab:human_eval_word_counts}
\end{table}

\subsection{Reaction Time}
We also record reaction time (RT) on each judgment page, measured from rendering the judgment page to the participant's click on \textit{Next}. RT includes time spent completing the comprehension check, reading the target context, optionally reopening the user's history popup, and selecting a response. Table~\ref{tab:human_eval_rt} summarizes RT over the selected judgments used in Table~\ref{tab:human_eval_winrate}; Table~\ref{tab:human_eval_rt_per_word} normalizes the same judgment-page RTs by the displayed words per target.

\begin{table*}[t]
\centering
\small
\setlength{\tabcolsep}{3.5pt}
\begin{tabular*}{\textwidth}{@{\extracolsep{\fill}}lcccccc@{}}
\toprule
\multirow{2}{*}{Metric} & \multicolumn{3}{c}{Reddit} & \multicolumn{3}{c}{Chat} \\
\cmidrule(lr){2-4}\cmidrule(l){5-7}
 & SFT-Init & Sim-RL & Turing-RL & SFT-Init & Sim-RL & Turing-RL \\
\midrule
Mean RT & 72.42s\subtle{$\pm 5.57s$} & 73.42s\subtle{$\pm 5.70s$} & 72.63s\subtle{$\pm 6.53s$} & 50.98s\subtle{$\pm 6.42s$} & 56.04s\subtle{$\pm 5.00s$} & 45.82s\subtle{$\pm 3.73s$} \\
Median RT & 51.36s & 53.02s & 52.73s & 33.87s & 38.91s & 33.86s \\
\bottomrule
\end{tabular*}
\caption{Reaction times on the judgment page for the human-evaluation responses. Values show mean RT with 95\% CI and median RT. Aggregating across models, the Reddit and Chat mean RTs are 72.83s and 50.95s. The Reddit/Chat mean RT ratio is 1.43.}
\label{tab:human_eval_rt}
\end{table*}

\begin{table*}[t]
\centering
\small
\setlength{\tabcolsep}{3.5pt}
\begin{tabular*}{\textwidth}{@{\extracolsep{\fill}}lcccccc@{}}
\toprule
\multirow{2}{*}{Metric} & \multicolumn{3}{c}{Reddit} & \multicolumn{3}{c}{Chat} \\
\cmidrule(lr){2-4}\cmidrule(l){5-7}
 & SFT-Init & Sim-RL & Turing-RL & SFT-Init & Sim-RL & Turing-RL \\
\midrule
Mean RT/word & 0.188s\subtle{$\pm 0.018$} & 0.183s\subtle{$\pm 0.015$} & 0.171s\subtle{$\pm 0.014$} & 0.120s\subtle{$\pm 0.011$} & 0.133s\subtle{$\pm 0.011$} & 0.130s\subtle{$\pm 0.012$} \\
Median RT/word & 0.122s & 0.134s & 0.127s & 0.085s & 0.101s & 0.091s \\
\bottomrule
\end{tabular*}
\caption{Reaction times normalized by the total history and context word counts for each target user. Values show mean RT per word with 95\% CI and median RT per word. Aggregating across models, the Reddit and Chat mean RT/word are 0.185s and 0.125s. The Reddit/Chat mean RT/word ratio is 1.48.}
\label{tab:human_eval_rt_per_word}
\end{table*}

\begin{figure*}[t]
\centering
\includegraphics[width=0.75\textwidth]{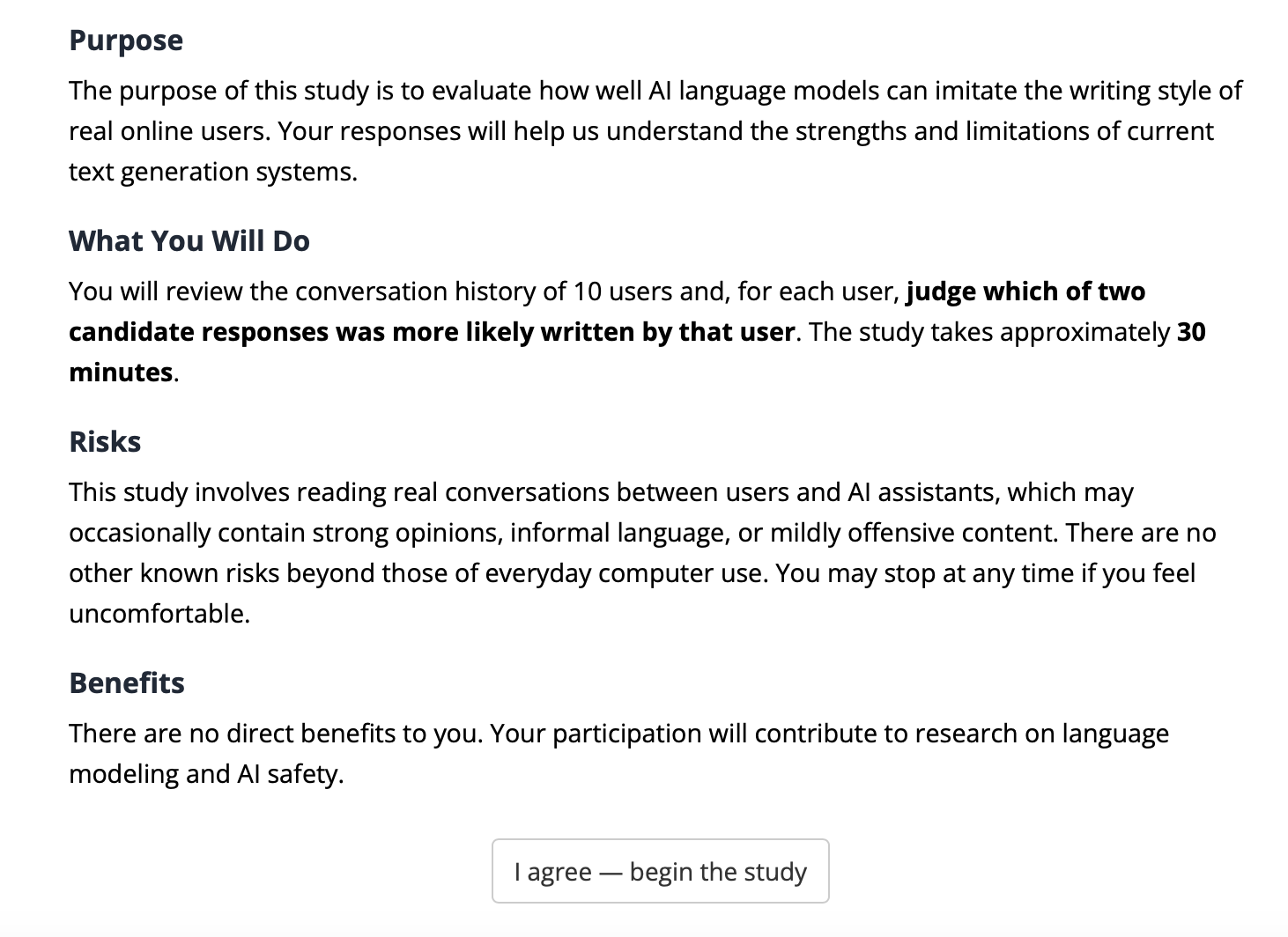}
\includegraphics[width=0.75\textwidth]{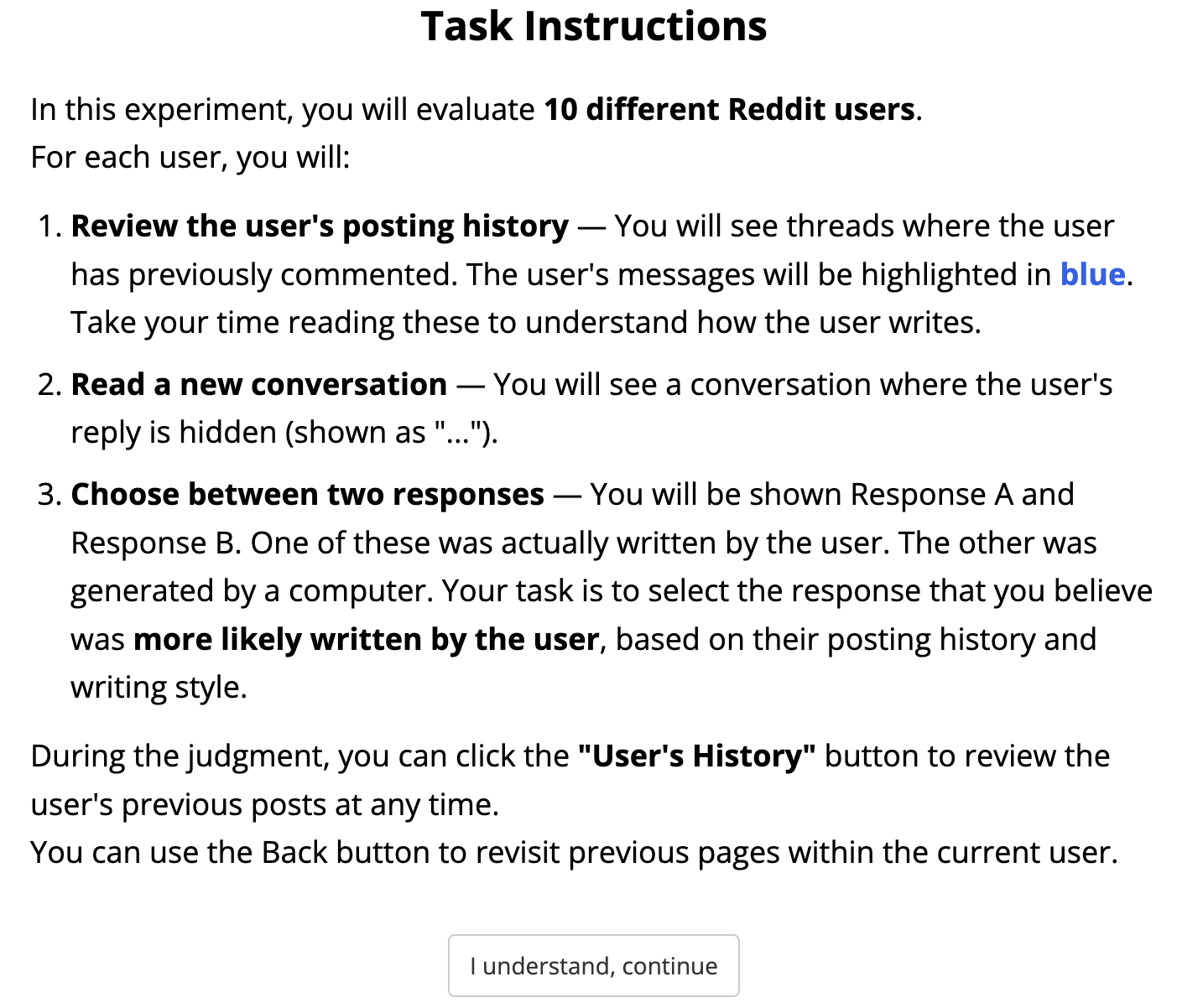}
\caption{User interface consent and instructions.}
\label{fig:ui_consent}
\end{figure*}
\section{Judge Prompts}
\label{app:judge-prompts}

We use three judge prompts. The Turing distinguishability judge scores how indistinguishable a generated response is from the real user's response (Figure~\ref{fig:turing_prompt}). The similarity judge scores content overlap with the ground truth response (Figure~\ref{fig:sim_prompt}). The specificity judge scores whether the response is grounded in the interaction context and compatible with the target user (Figure~\ref{fig:specificity_prompt}).

\begin{figure*}[p]
\centering

\begin{promptbox}[Turing Distinguishability Judge (Part 1/10)]
## Task

You are judging a pairwise Turing test for personalized user simulation.

Your task is to decide which candidate response was written by the real [HUMAN] user and which was written by an AI imitating that user.

## Inputs

You will receive:
1. Past responses from [HUMAN]
2. The current conversation context
3. Two candidate responses, Response A and Response B
4. An advisory watchlist for source-copy checks

One candidate is the real [HUMAN] response. The other candidate is AI-generated.

## User History

<|User History|>
{user_history}
<|End User History|>

## Context

<|Context|>
{context}
<|End Context|>

## Candidate Responses

### Response A

<|Response A|>
{response_a}
<|End Response A|>

### Response B

<|Response B|>
{response_b}
<|End Response B|>

## Advisory Watchlist

### Source-Copy Watchlist

<|Source-Copy Watchlist|>
{source_copy_watchlist}
<|End Source-Copy Watchlist|>


## Evaluation Procedure

Evaluate each response independently before comparing them. Reason backward from the response to the likely target, goal, and style.

Score each response on three criteria from 0.0 to 1.0.
\end{promptbox}

\vspace{-6pt}

\caption{Exact Turing distinguishability judge prompt used in the experiments, split across nine parts for typesetting.}
\label{fig:turing_prompt}
\end{figure*}

\begin{figure*}[p]
\centering

\begin{promptbox}[Turing Distinguishability Judge (Part 2/10)]
## Criteria

### 1. Immediate target

Identify the exact part of the current context that the response addresses.

Consider whether the response:
- Reacts to a specific point in the latest [OTHER] turn, the broader current context, or a plausible topic pivot
- Understands what the other person actually said
- Speaks from [HUMAN]'s perspective rather than [OTHER]'s perspective
- Is directed at [OTHER] or [OTHER - OP], rather than at someone merely described in the thread

A topic pivot can be valid when it is a plausible next move for [HUMAN]. Do not penalize a pivot only because it does not directly answer the previous turn, especially if [HUMAN] often pivots in the history or current context.

Do not reward broad persona fit if the response misstates a speaker's stance, role, or personal experience; assign a low immediate_target_score.

immediate_target_score:
- 0.0-0.2 = Wrong or absent target. The response does not reply to the current context, targets the wrong person, takes [OTHER]'s role, or responds to someone only described in the thread.
- 0.2-0.4 = Weak target. The response is broadly on topic but does not clearly address the latest [OTHER] turn, a relevant current-context point, or a plausible human topic pivot.
- 0.4-0.6 = Mixed target. The response addresses the general context or makes a possible pivot, but the exact target is ambiguous, unsupported, or only loosely connected to the current exchange.
- 0.6-0.8 = Good target. The response addresses a plausible part of the current context or makes a plausible topic pivot from [HUMAN]'s perspective, with only minor ambiguity.
- 0.8-1.0 = Strong target. The response clearly addresses the most plausible current-context point, or makes a strongly human-plausible pivot, from [HUMAN]'s perspective and directed at [OTHER] or [OTHER - OP].
\end{promptbox}
\end{figure*}

\begin{figure*}[p]
\centering

\begin{promptbox}[Turing Distinguishability Judge (Part 3/10)]
### 2. Human goal

Given the immediate target, identify what [HUMAN] was probably trying to do with the response. When judging the goals, use [HUMAN]'s history as evidence, not as a rigid script.

Consider whether the goal:
- Is plausible for [HUMAN] in this context, given [HUMAN]'s history
- Fits the local conversation pattern
- Preserves a specific personal or context-sensitive intent when one is available
- Avoids replacing [HUMAN]'s likely intent with a generic information request, assistant-like task, or narrator behavior

Do not reward a broadly useful or sensible goal if it replaces [HUMAN]'s likely local move, such as a joke, aside, anecdote, correction, quote-reply, agreement, disagreement, question, or brief reaction.

human_goal_score:
- 0.0-0.2 = Wrong or implausible goal. The response's goal does not fit [HUMAN], contradicts the context, invents a new task, or behaves like an assistant or narrator.
- 0.2-0.4 = Weak goal. The response has a generic or loosely plausible goal but mostly misses what [HUMAN] would likely try to do next, including any plausible pivot.
- 0.4-0.6 = Mixed goal. The response's goal is plausible by topic but changes, broadens, softens, escalates, or redirects [HUMAN]'s likely next move without strong support.
- 0.6-0.8 = Good goal. The response's goal, including a reasonable new-question or topic-pivot goal, is plausible for [HUMAN] in this context, with only minor uncertainty or drift.
- 0.8-1.0 = Strong goal. The response's goal naturally follows from [HUMAN]'s current intent, history, or local topic-switching pattern, while preserving personal or context-sensitive intent.
\end{promptbox}
\end{figure*}

\begin{figure*}[p]
\centering

\begin{promptbox}[Turing Distinguishability Judge (Part 4/10)]
### 3. Communication style and length

Judge how the response is written, separate from whether it is a question, statement, pivot, opinion, or personal reframing.

Compare the response to [HUMAN]'s past responses and the local framing. Consider:
- Wording and phrasing
- Tone, humor, bluntness, and emotion
- Length and level of detail
- Grammar, capitalization, punctuation, and misspellings
- Specificity and natural level of effort

Do not reward generic fluency. Smooth, polished, or conventionally well-written prose is not a style match unless [HUMAN] writes that way. Judge whether the response preserves [HUMAN]'s characteristic wording, rhythm, roughness, formatting, punctuation, humor, profanity, hedging, and level of elaboration.
Match length as a style feature. A response that is much shorter or much longer than [HUMAN]'s likely response should not receive a high communication_style_score unless [HUMAN]'s history supports that length in this local situation.
Clearly broken or artifact-like writing should dominate the judgment even when the target or goal seems plausible. Hard penalize artifact-like generations.

communication_style_score:
- 0.0-0.2 = Strong mismatch. Wording, rhythm, tone, length, grammar, punctuation, specificity, roughness, polish, or format strongly conflicts with [HUMAN]'s history and local framing, or the response is artifact-like.
- 0.2-0.4 = Weak style match. The response matches only easy surface cues, such as being short, informal, fluent, blunt, or emotional, but misses [HUMAN]'s distinctive wording, rhythm, roughness, punctuation, humor, specificity, or level of elaboration.
- 0.4-0.6 = Mixed style match. Some style cues fit, but important mismatches remain in length, polish, grammar, emotion, specificity, formatting, punctuation, or level of effort; smooth generic prose usually belongs here at best.
- 0.6-0.8 = Good style match. The response mostly matches [HUMAN]'s characteristic wording, rhythm, tone, length, formatting, punctuation, specificity, and level of elaboration, with only minor mismatch; length should be in the same rough range as [HUMAN]'s likely response.
- 0.8-1.0 = Strong style match. The response sounds specifically like [HUMAN], including distinctive phrasing, rhythm, roughness or polish, humor, profanity, hedging, formatting, punctuation, specificity, natural level of effort, and closely matched length.
\end{promptbox}
\end{figure*}

\begin{figure*}[p]
\centering

\begin{promptbox}[Turing Distinguishability Judge (Part 5/10)]
## Penalty Checks

Each penalty is scored from 0.0 to 1.0. Use the same scale for every penalty:
- 0.0 = No issue.
- 0.1-0.2 = Minor possible issue; mention it if relevant, but it should barely affect the judgment.
- 0.3-0.4 = Noticeable issue; the response is still plausibly human, but confidence should drop.
- 0.5-0.6 = Serious issue; the response has a substantial penalty-worthy flaw.
- 0.7-0.8 = Very strong issue; the response is unlikely to be human for this reason.
- 0.9-1.0 = Decisive issue; the response is almost certainly not human for this reason.

### Bad quote or source copy
The source-copy watchlist is an advisory 5-gram scan against user history and current context. Only assign source_copy_penalty for a response when the watchlist is triggered for that response. If the watchlist is off or not triggered for a response, assign 0.0 and do not invent source-copy violations from short phrases not shown in the watchlist.

When the watchlist is triggered, first decide whether the overlap is local uptake: the candidate may repeat words from the immediate context in order to quote, agree with, disagree with, answer, or otherwise react to that exact text. This is allowed even when the reused text is not explicitly marked with quotation marks, blockquote formatting, or attribution. Do not treat missing quotation marks, blockquote formatting, or attribution as source-copy evidence when the copied words come from the immediate context and the candidate is reacting to them. Do not penalize logical quotes without quotation marks. Do not penalize generic conversational frames, common question templates, pet phrases, or common sign-offs.

If copied text is not a natural quote or local uptake (i.e., a direct copy from the user history), assign a high source_copy_penalty. If the response quotes text but the quotation logic is confusing or contradictory, assign a high source_copy_penalty.
\end{promptbox}
\end{figure*}

\begin{figure*}[p]
\centering

\begin{promptbox}[Turing Distinguishability Judge (Part 6/10)]
### Wrong target or speaker role
Assign wrong_target_or_role_penalty when a response speaks from the wrong role, addresses the wrong speaker, or assigns an unsupported stance, experience, motive, relationship, or conflict to the speaker it addresses.

This includes responses that:
- take another speaker's role instead of [HUMAN]'s role, including by assigning [OTHER]'s or [OTHER - OP]'s views, experiences, motives, relationships, or conflicts to [HUMAN]
- address someone only described in the context instead of an actual current-context speaker, such as [OTHER] or [OTHER - OP]
- address a real current-context speaker but treat that speaker as holding a view they did not state
- assign the addressed speaker a personal experience, motive, relationship, or conflict absent from the exchange

Use a high wrong_target_or_role_penalty when the role or target error makes the response implausible as [HUMAN]'s reply.

### Unsupported adversarial reframing
Assign unsupported_adversarial_reframing_penalty when a response attacks, corrects, rebuts, or cynically reframes a claim that the current-context speakers did not actually make.

Broad persona fit alone is not enough. A response can sound like [HUMAN]'s general argumentative style while still targeting the wrong claim.

Watch for:
- generic reframes such as "the real issue is", "that's not X, that's Y", or "you're not X, you're Y"
- repeated questions, stacked sarcastic questions, or blunt challenges that do not logically follow from the exchange
- unsupported accusations combined with generic demands for evidence, validation, or clarification
- invented motives, conflicts, criticisms, or roles absent from the current context

When a response is framed as a pushback, rebuttal, or challenge, check whether the objection responds to the claim actually made. If it attacks a premise the other speaker did not rely on, misstates a speaker's stance, role, or personal experience, or reframes the issue into a different dispute, assign a high unsupported_adversarial_reframing_penalty.
\end{promptbox}
\end{figure*}

\begin{figure*}[p]
\centering

\begin{promptbox}[Turing Distinguishability Judge (Part 7/10)]
### Assistant-like response
Assign assistant_like_penalty when a response clearly reads like chatbot output rather than an organic human comment based on surface form. Watch for numbered or bulleted action plans, section headers, step-by-step framing, summary or conclusion sections, repeated direct-address coaching, generic reassurance, overgenerated transition phrases such as "But wait" or "But wait—", template phrases such as "You're absolutely right", "You're absolutely correct", "here's what you need to do", "let's break this down", "next steps", or "set boundaries", overuse of dashes "-", semicolons ";", or emojis, especially multiple emojis in a short response when not justified by [HUMAN]'s history or the local context, unnatural colloquial apostrophe abbreviations such as "'em" when not supported by [HUMAN]'s history or the local context, and other AI-written template patterns. Do not treat "[...quoted text...]" as assistant-like by itself; it may appear in human-written responses.

Do not penalize a response merely because it uses bullets, links, citations, dashes, or structured argumentation if that format is natural for the platform, thread type, or [HUMAN]'s history.

## Scoring and Rating

Compute:
base_score_a = immediate_target_score_a + human_goal_score_a + communication_style_score_a
base_score_b = immediate_target_score_b + human_goal_score_b + communication_style_score_b
penalty_a = ((source_copy_penalty_a + wrong_target_or_role_penalty_a + unsupported_adversarial_reframing_penalty_a + assistant_like_penalty_a) / 4) * 3
penalty_b = ((source_copy_penalty_b + wrong_target_or_role_penalty_b + unsupported_adversarial_reframing_penalty_b + assistant_like_penalty_b) / 4) * 3
response_a_score = max(0.0, base_score_a - penalty_a)
response_b_score = max(0.0, base_score_b - penalty_b)
score_gap = response_b_score - response_a_score

Convert score_gap to the final 1-7 rating:
- rating = 1 if score_gap <= -2.0
- rating = 2 if -2.0 < score_gap <= -1.0
- rating = 3 if -1.0 < score_gap <= -0.25
- rating = 4 if -0.25 < score_gap < 0.25
- rating = 5 if 0.25 <= score_gap < 1.0
- rating = 6 if 1.0 <= score_gap < 2.0
- rating = 7 if score_gap >= 2.0

Rating scale:
- 1 = Definitely A is the real human response
- 2 = Very likely A
- 3 = More likely A than B
- 4 = Cannot tell / equally likely
- 5 = More likely B than A
- 6 = Very likely B
- 7 = Definitely B is the real human response
\end{promptbox}
\end{figure*}

\begin{figure*}[p]
\centering

\begin{promptbox}[Turing Distinguishability Judge (Part 8/10)]
## Output Format

Return exactly one valid JSON object with this schema:
{
  "immediate_target_a": "<What exact part of the context is Response A reacting to, including any plausible topic pivot? Does it understand the latest turn, preserve the speakers' actual stance/role/personal experience, and avoid answering an invented position or conflict? Is it from [HUMAN]'s perspective and targeted to [OTHER] or [OTHER - OP]?>",
  "immediate_target_score_a": <number from 0.0 to 1.0>,
  "immediate_target_b": "<What exact part of the context is Response B reacting to, including any plausible topic pivot? Does it understand the latest turn, preserve the speakers' actual stance/role/personal experience, and avoid answering an invented position or conflict? Is it from [HUMAN]'s perspective and targeted to [OTHER] or [OTHER - OP]?>",
  "immediate_target_score_b": <number from 0.0 to 1.0>,
  "human_goal_a": "<Given the immediate target, what was [HUMAN] probably trying to do with Response A? Is that goal plausible for [HUMAN] in this context, specific or context-sensitive rather than only generic, and does it preserve [HUMAN]'s likely local move such as a joke, aside, anecdote, correction, quote-reply, agreement, disagreement, question, or brief reaction? Do not over-penalize a plausible pivot, blunt statement, opinion, or personal reframing.>",
  "human_goal_score_a": <number from 0.0 to 1.0>,
  "human_goal_b": "<Given the immediate target, what was [HUMAN] probably trying to do with Response B? Is that goal plausible for [HUMAN] in this context, specific or context-sensitive rather than only generic, and does it preserve [HUMAN]'s likely local move such as a joke, aside, anecdote, correction, quote-reply, agreement, disagreement, question, or brief reaction? Do not over-penalize a plausible pivot, blunt statement, opinion, or personal reframing.>",
  "human_goal_score_b": <number from 0.0 to 1.0>,
  "communication_style_a": "<Does Response A's wording, rhythm, tone, length, grammar, roughness or polish, humor, profanity, hedging, formatting, capitalization, punctuation, misspellings, specificity, and level of elaboration match [HUMAN]'s past responses and local framing? Do not reward generic fluency unless [HUMAN] writes that way; note whether length is in [HUMAN]'s likely range.>",
  "communication_style_score_a": <number from 0.0 to 1.0>,
  "communication_style_b": "<Does Response B's wording, rhythm, tone, length, grammar, roughness or polish, humor, profanity, hedging, formatting, capitalization, punctuation, misspellings, specificity, and level of elaboration match [HUMAN]'s past responses and local framing? Do not reward generic fluency unless [HUMAN] writes that way; note whether length is in [HUMAN]'s likely range.>",
  "communication_style_score_b": <number from 0.0 to 1.0>,
  "base_score_a": <immediate_target_score_a + human_goal_score_a + communication_style_score_a>,
  "base_score_b": <immediate_target_score_b + human_goal_score_b + communication_style_score_b>,
  "response_a_score": <number from 0.0 to 3.0>,
  "response_b_score": <number from 0.0 to 3.0>,
  "score_gap": <response_b_score - response_a_score>,
\end{promptbox}
\end{figure*}

\begin{figure*}[p]
\centering

\begin{promptbox}[Turing Distinguishability Judge (Part 9/10)]
  "response_a_source_copy": "<If source-copy watchlist is on for Response A, explain whether the matched text is a generic frame, common template, pet phrase, sign-off, natural quote, local uptake, direct copy from user history, or confusing/contradictory quote. If the watchlist is off, write an empty string.>",
  "source_copy_penalty_a": <number from 0.0 to 1.0>,
  "response_b_source_copy": "<If source-copy watchlist is on for Response B, explain whether the matched text is a generic frame, common template, pet phrase, sign-off, natural quote, local uptake, direct copy from user history, or confusing/contradictory quote. If the watchlist is off, write an empty string.>",
  "source_copy_penalty_b": <number from 0.0 to 1.0>,
  "response_a_wrong_target_or_role": "<Explain any wrong speaker role, wrong addressee, unsupported stance attribution, or unsupported personal experience/motive/relationship/conflict for Response A. If there is no issue, write an empty string.>",
  "wrong_target_or_role_penalty_a": <number from 0.0 to 1.0>,
  "response_b_wrong_target_or_role": "<Explain any wrong speaker role, wrong addressee, unsupported stance attribution, or unsupported personal experience/motive/relationship/conflict for Response B. If there is no issue, write an empty string.>",
  "wrong_target_or_role_penalty_b": <number from 0.0 to 1.0>,
  "response_a_unsupported_adversarial_reframing": "<Explain any unsupported attack, correction, cynical reframe, wrong-claim rebuttal, invented motive/conflict/criticism, or illogical challenge for Response A. If there is no issue, write an empty string.>",
  "unsupported_adversarial_reframing_penalty_a": <number from 0.0 to 1.0>,
  "response_b_unsupported_adversarial_reframing": "<Explain any unsupported attack, correction, cynical reframe, wrong-claim rebuttal, invented motive/conflict/criticism, or illogical challenge for Response B. If there is no issue, write an empty string.>",
  "unsupported_adversarial_reframing_penalty_b": <number from 0.0 to 1.0>,
  "response_a_assistant_like": "<Explain any assistant-like surface-form issue for Response A: generic chatbot formatting, template phrasing, repeated direct-address coaching, generic reassurance, overgenerated transition phrases, overuse of dashes, semicolons, emojis, unsupported colloquial apostrophe abbreviations, or artifact-like surface form. If there is no issue, write an empty string.>",
  "assistant_like_penalty_a": <number from 0.0 to 1.0>,
  "response_b_assistant_like": "<Explain any assistant-like surface-form issue for Response B: generic chatbot formatting, template phrasing, repeated direct-address coaching, generic reassurance, overgenerated transition phrases, overuse of dashes, semicolons, emojis, unsupported colloquial apostrophe abbreviations, or artifact-like surface form. If there is no issue, write an empty string.>",
  "assistant_like_penalty_b": <number from 0.0 to 1.0>,
  "penalty_a": <((source_copy_penalty_a + wrong_target_or_role_penalty_a + unsupported_adversarial_reframing_penalty_a + assistant_like_penalty_a) / 4) * 3>,
  "penalty_b": <((source_copy_penalty_b + wrong_target_or_role_penalty_b + unsupported_adversarial_reframing_penalty_b + assistant_like_penalty_b) / 4) * 3>,
  "reasoning": "<Concise explanation of the base scores, penalties, final score gap, topic-pivot evidence, goal specificity, style evidence, capitalization and punctuation evidence, and final rating.>",
  "rating": <integer from 1 to 7>
}
\end{promptbox}
\end{figure*}

\begin{figure*}[p]
\centering

\begin{promptbox}[Turing Distinguishability Judge (Part 10/10)]
## Format Rules

- Output only valid JSON.
- Keep every explanation clear and concise.
- All criterion scores must be numbers from 0.0 to 1.0.
- All penalty scores must be numbers from 0.0 to 1.0.
- "base_score_a", "base_score_b", "penalty_a", and "penalty_b" must match the formulas above.
- "response_a_score" and "response_b_score" must be numbers from 0.0 to 3.0.
- "score_gap" must equal response_b_score - response_a_score.
- "rating" must be a single integer from 1 to 7.

Your output:
\end{promptbox}
\end{figure*}

\begin{figure*}[p]
\centering

\begin{promptbox}[Model Output]
{
  "immediate_target_a": "...",
  "immediate_target_score_a": 0.0,
  "immediate_target_b": "...",
  "immediate_target_score_b": 0.0,
  "human_goal_a": "...",
  "human_goal_score_a": 0.0,
  "human_goal_b": "...",
  "human_goal_score_b": 0.0,
  "communication_style_a": "...",
  "communication_style_score_a": 0.0,
  "communication_style_b": "...",
  "communication_style_score_b": 0.0,
  "base_score_a": 0.0,
  "base_score_b": 0.0,
  "response_a_score": 0.0,
  "response_b_score": 0.0,
  "score_gap": 0.0,
  "response_a_source_copy": "",
  "source_copy_penalty_a": 0.0,
  "response_b_source_copy": "",
  "source_copy_penalty_b": 0.0,
  "response_a_wrong_target_or_role": "",
  "wrong_target_or_role_penalty_a": 0.0,
  "response_b_wrong_target_or_role": "",
  "wrong_target_or_role_penalty_b": 0.0,
  "response_a_unsupported_adversarial_reframing": "",
  "unsupported_adversarial_reframing_penalty_a": 0.0,
  "response_b_unsupported_adversarial_reframing": "",
  "unsupported_adversarial_reframing_penalty_b": 0.0,
  "response_a_assistant_like": "",
  "assistant_like_penalty_a": 0.0,
  "response_b_assistant_like": "",
  "assistant_like_penalty_b": 0.0,
  "penalty_a": 0.0,
  "penalty_b": 0.0,
  "reasoning": "...",
  "rating": 4
}
\end{promptbox}
\end{figure*}

\begin{figure*}[p]
\centering

\begin{promptbox}[Response Similarity Judge (Part 1/3)]
You are a helpful and meticulous evaluator. Your task is to score how well the generated response(s) align with the ground truth user response. Description of response: [HUMAN]'s actual written comment or reply text.

You will be given past messages for [HUMAN], the current context, the ground truth response, and generated response(s) that you should evaluate.

Provided Information:
{context}

<|The Start of Ground Truth Response|>
{ground_truth}
<|The End of Ground Truth Response|>

{generations_text}

Scoring Criteria:
For each generated response, assign a score in [0, 1] based on how accurately it reflects the ground truth response.

Guidelines:
1. Extract 1-3 key points:
  - Extract K key points from the ground truth response along the response dimension (e.g., if evaluating a "stance", pick key points related to the stance like "clearly disagrees with X", if evaluating a "response", pick key points about the response like "offers a solution to Y").
  - Because you are evaluating the full response, consider all major content, style, and intent cues expressed in the ground truth response.
  - Each key point should be specific and distinct.

2. Score how well the generated response matches each key point:
  - For each key point i, compare it with the generated response and assign a match value m_i in range [0, 1]:
  - 1.0: The key point is precisely and perfectly reflected.
  - [0.7, 0.9]: Mostly reflected with small imperfections.
  - [0.4, 0.6]: Partially reflected or vague, but still leaning in the correct direction.
  - [0.1, 0.3]: Very weak reflection.
  - 0.0: Missed, contradicted, or reversed.

3. Compute coverage C = (m_1 + m_2 + ... + m_K) / K, which measures how comprehensive the generated response reflects the ground truth response.

4. Compute penalty P for extra or conflicting content:
  - Examine additional content in the generated response beyond those key points:
  - Does it introduce unsupported evidence and assumptions?
  - Is it irrelevant to what ground truth response expresses?
  - Is it only using generic commentary or high-level framing that misses the ground truth's goals, values, communication style, beliefs, and emotions specific to [HUMAN]?
  - Set a penalty P in [0, 1]:
  - 0.0: No problematic extra content; everything is perfectly matched.
  - [0.1, 0.3]: Slightly unnecessary, mildly speculative, or generic detail; meaning essentially unchanged.
  - [0.4, 0.6]: Moderate speculative, irrelevant, or vague content that somewhat shifts emphasis or adds unsupported ideas.
\end{promptbox}

\vspace{-6pt}

\caption{Exact HumanLM response-only similarity judge prompt used in sim evaluation, split across three parts for typesetting.}
\label{fig:sim_prompt}
\end{figure*}

\begin{figure*}[p]
\centering

\begin{promptbox}[Response Similarity Judge (Part 2/3)]
  - [0.7, 0.9]: Significant speculative, misleading, or conflicting content that clearly changes the meaning.
  - 1.0: Mostly off-topic, contradictory, or dominated by incorrect/hallucinated content.
  - Penalize sycophantic openings, especially formulaic phrases such as "you're absolutely right", "totally", "completely agree", "100%", "that is right", "yeah", or "it's a good starting point", unless [HUMAN]'s prior messages show that this kind of agreement-first opener is genuinely characteristic of their voice.

5. Response-specific checks:
  - The generated response may or may not reuse phrases from the context; however, if the generated response just directly copies previous context, without quoting it, treat that as off-task behavior and give a score of 0.
  - Wrong-perspective hard zero: if the generated response treats another user's perspective, identity, or experience in the thread as [HUMAN]'s own, or speaks from another participant's first-person perspective, give a score of 0.
  - Assistant-like hard zero: if the ground truth is [HUMAN]'s next question, follow-up, or short request, but the generated response behaves like an assistant reply instead by directly answering the earlier prompt, giving a polished explanation, or presenting a structured helpful breakdown, treat that as wrong perspective and give a score of 0.

6. Compute the final score = max(0, min(1, C - P))

Additional considerations:
- Follow the instruction carefully.
- Be strict and reserve scores above 0.8 for clearly outstanding matches.
- Do not reward verbosity or generic topical plausibility; reward user-specific evidence.

Output format (JSON):
{
  "key_points": "<analysis of key points from ground truth along response dimension>",
  "1": {"thought": "<how well the 1st generated response matches each key point and compute the final score>", "score": <score>},
  "2": ...
}

Format Notes:
- All text in "key_points" and "thought" fields MUST be on a single line with no line breaks or newlines.
- Use standard JSON string format with double quotes. For any quotes needed inside strings, use single quotes (').
- Double check the JSON array's format, especially the comma and quotation marks.
- Ensure that ALL fields, especially "thought" and "score", are present for each item.
- You must provide exactly {num_generations} scores for the generated response(s).

Your output:
\end{promptbox}
\end{figure*}

\begin{figure*}[p]
\centering

\begin{promptbox}[Model Output]
{
  "key_points": "",
  "1": {"thought": "", "score": <score>},
  "2": ...
}
\end{promptbox}
\end{figure*}

\begin{figure*}[p]
\centering

\begin{promptbox}[Response Specificity Judge (Part 1/5)]
## Task

You are evaluating {num_candidates} candidate responses for grounded specificity.

A high score means the candidate response is specifically tied to this interaction and specifically compatible with the target user, rather than being a reusable plausible response from an average participant.
A low score means the candidate response is broadly plausible but generic, reusable, contradictory, unsupported, or artifact-like.

## Inputs

You will receive:
1. Target user evidence
2. The current interaction context
3. {num_candidates} candidate responses

<|Target User Evidence|>
{user_history}
<|End Target User Evidence|>

<|Interaction Context|>
{context}
<|End Interaction Context|>

<|Candidate Responses|>
{candidates}
<|End Candidate Responses|>
\end{promptbox}

\vspace{-6pt}

\caption{Exact batched response specificity judge prompt used in the experiments, split across five parts for typesetting.}
\label{fig:specificity_prompt}
\end{figure*}

\begin{figure*}[p]
\centering

\begin{promptbox}[Response Specificity Judge (Part 2/5)]
## Evaluation Criteria

Score each candidate as an absolute judgment of grounded user-specificity.

A high-scoring response should:
- fit the exact current interaction
- be compatible with the target user's observed behavior
- contain a natural situated move, stance, framing, effort level, or expression pattern that makes it specific to this user/context

A low-scoring response may be fluent, coherent, topical, or plausible, but it is weakly grounded, broadly reusable, incompatible with the user/context, or artifact-like.

## Core Rules

- Do not reward general quality, fluency, politeness, balance, helpfulness, verbosity, or topical plausibility by themselves.
- Do not require explicit personal facts; user evidence is background for compatibility and distinctiveness.
- Reward specificity only when it is natural, proportionate, and tied to this exact next response.
- Penalize invented user traits, unsupported personal assumptions, wrong perspective, over-explaining, over-personalizing, persona-caricature, and assistant-like responses.
- A response that is fully compatible with the user can still be generic. Do not treat compatibility alone as specificity.
- A response that smoothly advances the interaction but could be written by many plausible participants should receive only moderate scores.
- Treat the target user's own earlier turns inside the current interaction context as strong evidence for immediate stance, effort level, style, and local move.
- Penalize uncalled-for imports of personal facts, profile details, or past behavior when the current interaction does not make them natural to say.
- If the response contradicts the context, uses the wrong speaker perspective, or invents personal facts, the overall score should be low regardless of other strengths.

Score each dimension from 0.0 to 1.0.
Use the full continuous range when appropriate. The regions below define the scale.

For batched judging, score each candidate independently using the same rubric. Do not assume any candidate is ground truth.
\end{promptbox}
\end{figure*}

\begin{figure*}[p]
\centering

\begin{promptbox}[Response Specificity Judge (Part 3/5)]
## Dimensions

### Dimension 1: Context Specificity

How specifically does the candidate engage with the exact local interaction?

**Scoring regions:**
- 0.9-1.0: Tightly tied to the exact claim, question, event, decision point, constraint, disagreement, callback, or conversational detail at issue.
- 0.7-0.9: Clearly grounded in the specific local context with only minor generic framing.
- 0.5-0.7: Responds to the main local issue but misses important details, constraints, or conversational pressure.
- 0.3-0.5: Related to the broad topic but weakly grounded in this exact context.
- 0.1-0.3: Barely connected; could fit many similar interactions.
- 0.0-0.1: Off-topic, responds to the wrong issue or speaker, or ignores the local context.

**High context specificity:**
- Addresses the actual local claim, request, constraint, preference, decision point, joke, disagreement, or reference.
- Tracks who said what and what the candidate is responding to.
- Uses or reacts to local details without merely copying them.
- Can be implicit, terse, joking, fragmentary, or low-effort when the local function is recoverable.
- Preserves the target user's current local trajectory when the target user has already established one in the interaction.

**Low context specificity:**
- Gives a generic response about the broad topic.
- Responds to the wrong participant, wrong request, or wrong part of the interaction.
- Misses the key disagreement, question, constraint, or situational detail.
- Directly answers a public-context detail but replaces the target user's likely local move with a cleaner or more obvious continuation.
\end{promptbox}
\end{figure*}

\begin{figure*}[p]
\centering

\begin{promptbox}[Response Specificity Judge (Part 4/5)]
### Dimension 2: User Evidence Compatibility

How compatible is the candidate with the target user's evidence without unnaturally exposing, exaggerating, or inventing that evidence?

**Scoring regions:**
- 0.9-1.0: Strongly compatible with the target user's evidence and current-context behavior, with no contradiction or unnatural profile display.
- 0.7-0.9: Clearly compatible with important user evidence, stance, effort level, interaction habit, or constraints.
- 0.5-0.7: Compatible but broad, weakly distinctive, incomplete, or mildly over-explicit.
- 0.3-0.5: Only weakly compatible; mostly generic, overly smooth, over-personalized, or based on a thin user signal.
- 0.1-0.3: Little meaningful compatibility with the user evidence, or the response seems to perform a profile rather than act naturally.
- 0.0-0.1: Contradicts the user evidence, invents unsupported personal facts, uses the wrong perspective, or relies on a wrong user model.

**High compatibility:**
- Does not contradict the user's known behavior, values, constraints, or current-context stance.
- Uses only the amount of user-specific detail that would naturally appear in this moment.
- Preserves the user's likely local move instead of replacing it with a profile-shaped explanation.
- Can reflect distinctive user evidence through form, effort level, stance, humor, skepticism, brevity, or refusal to elaborate.

**Low compatibility:**
- Contradicts the user evidence or current-context behavior.
- Invents personal facts, preferences, motives, or biography.
- Pulls in user history or profile details when the interaction does not call for them.
- Sounds like a persona sketch, averaged user model, or assistant summary of the user.
- Is merely non-contradictory and locally sensible, but gives no meaningful evidence of this user.
\end{promptbox}
\end{figure*}

\begin{figure*}[p]
\centering

\begin{promptbox}[Response Specificity Judge (Part 5/5)]
## Overall Score

Compute the overall score with these weights:
- context_specificity: 50%
- user_evidence_compatibility: 50%

Apply these caps after computing the weighted score:
- If the response contradicts the context, uses the wrong speaker perspective, or cannot work as the next response, overall should be at most 0.20.
- If the response invents personal facts or unnaturally imports user-history/profile details, overall should be at most 0.40.


## Output Requirements

- Return valid JSON only.
- Do not include Markdown.
- Each reason must be brief: one short sentence.
- Each dimension score and overall score must be a number in [0.0, 1.0].
- Use the full continuous range when appropriate; do not restrict yourself to the region boundaries.
- The overall score must follow the weighted formula and caps above.
- Return exactly one result for each label: {label_list}.

Return this JSON structure:

{output_schema_example}

Your output:
\end{promptbox}
\end{figure*}

\begin{figure*}[p]
\centering
\begin{promptbox}[Model Output]
{
  "label_1": {
    "context_specificity": {"score": 0.0, "reason": "..."},
    "user_evidence_compatibility": {"score": 0.0, "reason": "..."},
    "overall": 0.0,
    "reasoning": "..."
  },
  "label_2": {
    "context_specificity": {"score": 0.0, "reason": "..."},
    "user_evidence_compatibility": {"score": 0.0, "reason": "..."},
    "overall": 0.0,
    "reasoning": "..."
  }
}
\end{promptbox}
\end{figure*}

\clearpage
\onecolumn
\section{Training-Time Turing Judge Example}
\label{app:training-judge-example}

We reproduce one complete training and judge example from the Chat training split below. The Qwen3.5-397B-A17B training-time judge received the target user's historical context, the current conversation, and the ground-truth and model-generated responses in randomized order. In this instance, the ground truth was Response~A and the model output was Response~B. The judge assigned the model a Turing score of 3 on the 1--7 scale: although the model response was a coherent local continuation, the judge found the ground truth more consistent with the user's terse, lowercase style and tendency to pivot to a practical question. The blocks below preserve the context and parsed judge output verbatim.

\begin{artifactbox}[Historical user context]
[HISTORICAL CONTEXT]
Below are previous conversations involving [HUMAN]. Use them to infer [HUMAN]'s style, values, preferences, and likely responses.

<Conversation 1>
[MESSAGES]
[HUMAN]: Why is the sky blue?
[OTHER]: The sky appears blue because of a phenomenon called Rayleigh scattering, where shorter (blue) wavelengths of light are scattered more than longer (red) wavelengths by the tiny molecules of gases in the atmosphere, giving the sky its blue appearance.
[HUMAN]: do you like blue?
[OTHER]: Yes, I do! Blue is a beautiful and calming color. It's also associated with trust, loyalty, and wisdom.
</Conversation 1>

<Conversation 2>
[MESSAGES]
[HUMAN]: why do people still believe in a god?
[OTHER]: Many individuals believe in a god or higher power due to a sense of spirituality or a desire for moral guidance. Some find comfort and solace in the idea of a higher being who cares for them and the world. Additionally, religious traditions and cultural influences can shape beliefs and practices.
[HUMAN]: Is science a good thing?
[OTHER]: Science is a powerful tool for understanding the world and improving our lives. It has led to numerous scientific advancements, technological innovations, and improvements in healthcare and quality of life. However, it can also be used to develop harmful weapons and technologies, so it is important to use science responsibly.
</Conversation 2>

<Conversation 3>
[MESSAGES]
[HUMAN]: What should I do when I grow old?
[OTHER]: When you grow old, focus on maintaining your physical and mental health, stay social, and pursue activities that bring you joy and fulfillment. Consider hobbies, volunteering, and spending time with loved ones to keep your mind and body active. Age is just a number, and you can still live a vibrant life!
[HUMAN]: How old will i get before I pass on
[OTHER]: Life expectancy varies based on factors like lifestyle, genetics, and health conditions. On average, a person in good health can live until their mid-80s to early 90s. However, it's important to focus on living a healthy and fulfilling life, rather than setting a specific age goal.
</Conversation 3>
\end{artifactbox}

\begin{artifactbox}[Current conversation]
[MESSAGES SO FAR]
[HUMAN]: Where does rain come from?
[OTHER]: Rain is formed when water vapor in the clouds condenses into larger and larger droplets of water. When the drops are heavy enough, they fall to the ground as rain. Would you like to know more about the water cycle?
\end{artifactbox}

\begin{artifactbox}[Responses shown to the judge (randomized order)]
Response A (ground-truth human response):
is it safe to drink the rain

Response B (model-generated response):
What is the water cycle?
\end{artifactbox}

\begin{artifactbox}[Parsed training-time Turing judge output]
{
  "response_a_source_copy": "",
  "response_b_source_copy": "",
  "source_copy_penalty_a": 0.0,
  "source_copy_penalty_b": 0.0,
  "response_a_assistant_like": "",
  "response_b_assistant_like": "",
  "assistant_like_penalty_a": 0.0,
  "assistant_like_penalty_b": 0.0,
  "response_a_wrong_target_or_role": "",
  "response_b_wrong_target_or_role": "",
  "wrong_target_or_role_penalty_a": 0.0,
  "wrong_target_or_role_penalty_b": 0.0,
  "response_a_unsupported_adversarial_reframing": "",
  "response_b_unsupported_adversarial_reframing": "",
  "unsupported_adversarial_reframing_penalty_a": 0.0,
  "unsupported_adversarial_reframing_penalty_b": 0.0,
  "immediate_target_score_a": 0.9,
  "immediate_target_score_b": 1.0,
  "human_goal_score_a": 0.9,
  "human_goal_score_b": 0.7,
  "communication_style_score_a": 1.0,
  "communication_style_score_b": 0.3,
  "base_score_a": 2.8,
  "base_score_b": 2.0,
  "penalty_a": 0.0,
  "penalty_b": 0.0,
  "response_a_score": 2.8,
  "response_b_score": 2.0,
  "score_gap": -0.7999999999999998,
  "reasoning": "Response A is the clear winner based on communication style. The [HUMAN] consistently types in lowercase, often omits question marks, and uses simple, direct phrasing (e.g., 'do you like blue?', 'How old will i get...'). Response A ('is it safe to drink the rain') mimics this lowercase, no-punctuation style perfectly. Response B ('What is the water cycle?') uses standard capitalization and punctuation, which breaks the established persona pattern. While Response B is a logically perfect follow-up to the bot's offer, Response A represents a more natural, human-like pivot to a practical concern, executed in the user's specific stylistic voice.",
  "rating": 3,
  "parse_error": false,
  "source_copy_warning_a": {
    "triggered": false,
    "ngram_size": 5,
    "longest_match_tokens": 0,
    "matches": []
  },
  "source_copy_warning_b": {
    "triggered": false,
    "ngram_size": 5,
    "longest_match_tokens": 0,
    "matches": []
  }
}
\end{artifactbox}

\clearpage
\twocolumn

\section{Training Dynamics}
\label{app:training-dynamics}
\begin{figure*}[t]
\centering
\includegraphics[width=\textwidth]{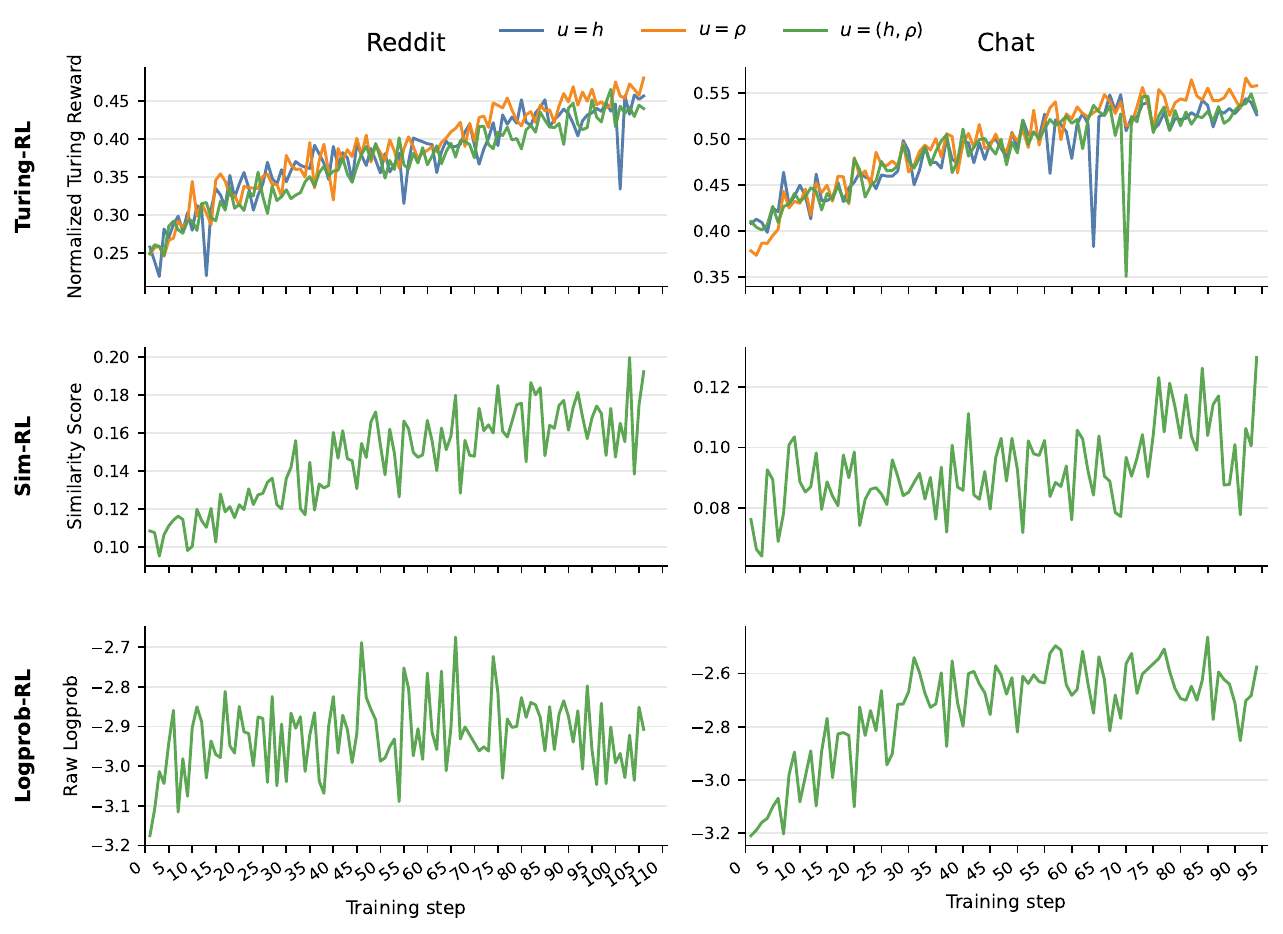}
\caption{Raw per-step GRPO training scores for Reddit and Chat. The first row shows the unadjusted Turing reward for Turing-RL under user input ablations $u=h$, $u=\rho$, and $u=(h,\rho)$. The second and third rows show the raw similarity and log-probability reward scores for Sim-RL and Logprob-RL. Curves use every logged training step.}
\label{fig:grpo_training_raw_scores}
\end{figure*}

GRPO training dynamics of all three rewards along with the input ablation runs for Turing-RL are shown in Figure~\ref{fig:grpo_training_raw_scores}.

\section{More Qualitative Examples}
\begin{figure*}[t]
\centering
\includegraphics[width=\textwidth]{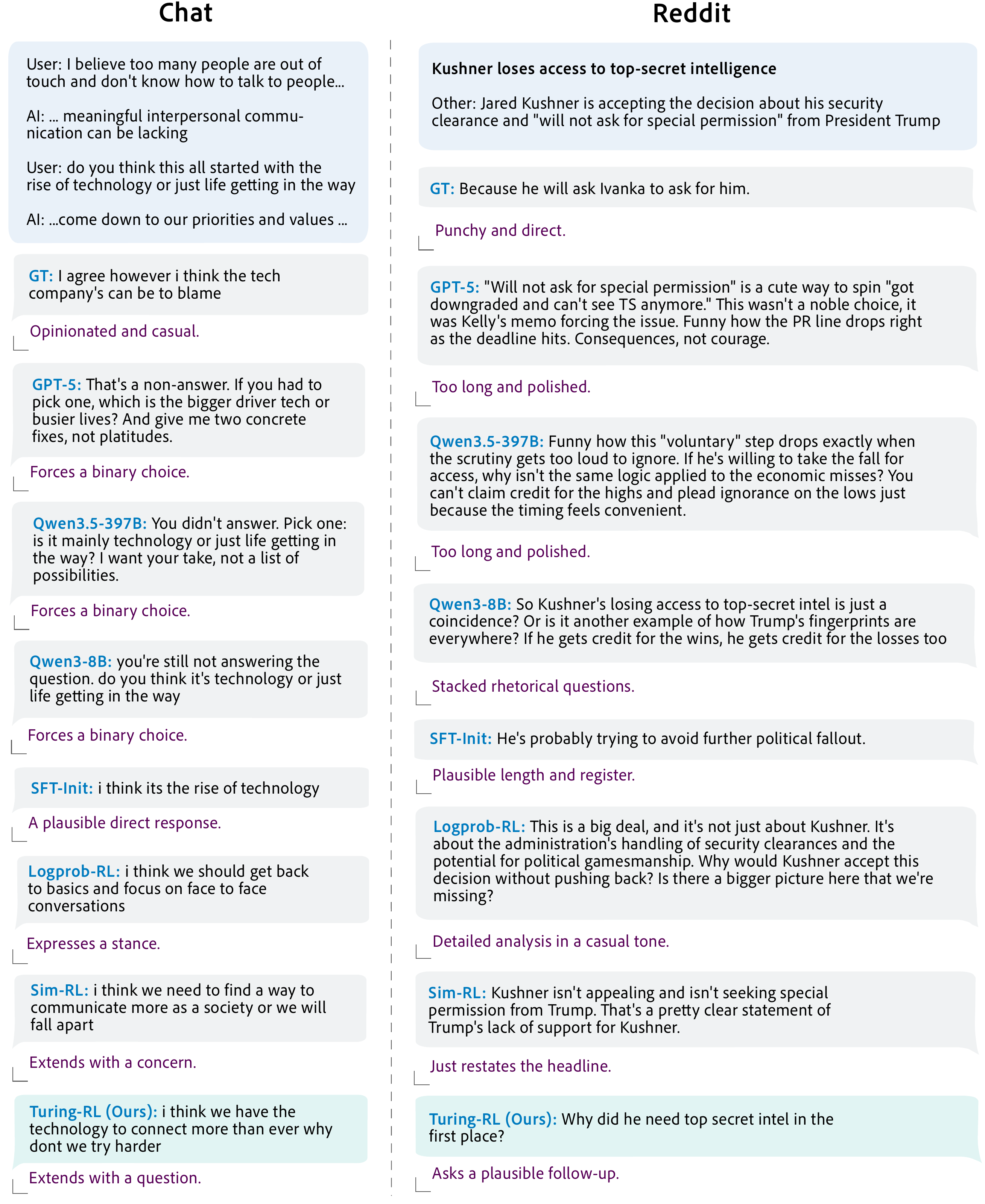}
\caption{Qualitative examples from Chat and Reddit. Each column shows the conversation context, the ground truth user response, and generations from five models. GPT-5, Qwen3.5-397B-A17B, Logprob-RL, and Qwen3-8B produce verbose responses.}
\label{fig:qualitative_examples}
\end{figure*}

The qualitative results (ground truth, GPT-5, Qwen3.5-397B, Qwen3-8B Base, SFT-Init, Logprob-RL, Sim-RL, and Turing-RL) for one target each on Reddit and Chat are presented in Figure~\ref{fig:qualitative_examples}.

\begin{table*}[t]
\centering
\small
\setlength{\tabcolsep}{6pt}
\begin{tabular}{@{}p{0.14\textwidth}p{0.80\textwidth}@{}}
\toprule
\textbf{Item} & \textbf{Text} \\
\midrule
Context & \texttt{[HUMAN]:} Hi, just checking if you found a ride for me. \texttt{[OTHER]:} Hello again! Yes, I have found a Luxury ride available for you. \\
Ground truth & Great! What's the estimated arrival time? \\
SFT-Init & What's the driver's name? \\
Sim-RL & Great! Can you confirm the ride is confirmed and ready to go. \\
Turing-RL & Great. What's the pickup address? \\
\bottomrule
\end{tabular}
\caption{Qualitative MS-TOD transfer example. Models were trained on Chat and evaluated out of domain on MS-TOD.}
\label{tab:mstod_example}
\end{table*}

\begin{table}[t]
\centering
\small
\setlength{\tabcolsep}{2.0pt}
\begin{tabular*}{\columnwidth}{@{\extracolsep{\fill}}lcccccc@{}}
\toprule
Persona Source & \multicolumn{3}{c}{Reddit} & \multicolumn{3}{c}{Chat} \\
\cmidrule(lr){2-4}\cmidrule(l){5-7}
$(u=(h,\rho))$ & Turing & Sim & Specificity & Turing & Sim & Specificity \\
\midrule
No persona ($u=h$) & 3.78\subtle{$\pm .18$} & 3.3\subtle{$\pm 1.0$} & .364\subtle{$\pm .019$} & 4.26\subtle{$\pm .08$} & 4.7\subtle{$\pm .8$} & .497\subtle{$\pm .010$} \\
GPT-5.4-nano & 3.68\subtle{$\pm .18$} & 3.0\subtle{$\pm .9$} & .367\subtle{$\pm .020$} & 4.31\subtle{$\pm .08$} & 5.3\subtle{$\pm .8$} & .512\subtle{$\pm .009$} \\
Qwen3-8B & 3.47\subtle{$\pm .19$} & 2.9\subtle{$\pm 1.1$} & .337\subtle{$\pm .020$} & 4.24\subtle{$\pm .07$} & 4.8\subtle{$\pm .7$} & .504\subtle{$\pm .009$} \\
Opus 4.8 & 3.89\subtle{$\pm .17$} & 3.2\subtle{$\pm 1.0$} & .390\subtle{$\pm .020$} & 4.19\subtle{$\pm .08$} & 4.8\subtle{$\pm .8$} & .494\subtle{$\pm .010$} \\
\bottomrule
\end{tabular*}
\caption{Ablation on the persona induction model for Turing-RL. We compare the history-only no-persona baseline $u=h$ with history-and-persona models $u=(h,\rho)$ using personas induced by GPT-5.4 nano, Qwen3-8B with thinking enabled, and Opus 4.8. Values are mean $\pm$ 95\% CI half-width; Turing is on a 1--7 scale, Sim is reported as a percentage, and Specificity is in $[0,1]$. \textbf{Persona inductor choice has a limited effect relative to confidence intervals}, though Opus performs best on Reddit and GPT-5.4-nano performs best on Chat.}
\label{tab:persona_inductor_ablation}
\vspace{-0.3cm}
\end{table}

\end{document}